\documentclass{article}

\usepackage[usenames,dvipsnames]{xcolor}
\usepackage[preprint]{corl_2026} % Use this for the initial submission.
\usepackage{graphicx}
\usepackage{wrapfig}
\usepackage{tikz}
\usetikzlibrary{arrows.meta,calc,fit,positioning}
\usepackage{array}
\usepackage{booktabs}
\usepackage{amsmath,amssymb}
\usepackage{pifont}
\usepackage{enumitem}
\usepackage{multirow}
\newcommand{\tightsection}[1]{\vspace{-0.9em}\section{#1}\vspace{-0.3em}}
\newcommand{\tightsubsection}[1]{\vspace{-0.8em}\subsection{#1}\vspace{-0.35em}}
\newcommand{\tightfloatbefore}{\vspace{-0.9em}}
\newcommand{\tightfloatafter}{\vspace{-0.9em}}

\title{ZETA: A Controlled Study of Zero-Shot Cross-Embodiment VLA Transfer for Tabletop Manipulation}

\author{
  \hspace{-1cm}
  \textbf{Mi Yan}$^{1,2}$\thanks{Equal contribution.} \hspace{0.2em}
  \textbf{Wenhao Zhang}$^{2,3}$\footnotemark[1] \hspace{0.2em}
  \textbf{Zhiqi Zhang}$^{2,3}$\footnotemark[1] \hspace{0.2em}
  \textbf{Yu Peng}$^{2,4}$\footnotemark[1] \hspace{0.2em}
  \textbf{Tangxinyu Wang}$^{2,3}$\footnotemark[1]
  \\[0.2em]
  \hspace{-1cm}
  \textbf{Lingfei Zhai}$^{2,3}$ \hspace{0.2em}
  \textbf{Jiayi Su}$^{2,6}$ \hspace{0.2em}
  \textbf{Shengliang Deng}$^{2,5}$ \hspace{0.2em}
  \textbf{Lin Peng}$^{2,7}$ \hspace{0.2em}
  \textbf{Yaowei Liu}$^{2,3}$ \hspace{0.2em}
  \textbf{Yuxing Chen}$^{1,2}$
  \\[0.2em]
  \hspace{-1cm}
  \textbf{Zhiyuan Wei}$^{2,3}$ \hspace{0.2em}
  \textbf{Jilong Wang}$^{2}$ \hspace{0.2em}
  \textbf{Jiayi Chen}$^{1,2}$ \hspace{0.2em}
  \textbf{Jiangran Lyu}$^{1,2}$ \hspace{0.2em}
  \textbf{Zhizheng Zhang}$^{2}$\thanks{
    Corresponding authors. Correspondence to
    \href{mailto:zhangzz@galbot.com}{zhangzz@galbot.com},
    \href{mailto:hewang@pku.edu.cn}{hewang@pku.edu.cn}.
  } \hspace{0.2em}
  \textbf{He Wang}$^{1,2}$\footnotemark[2]
  \\[0.5em]
  \hspace{-1cm}
  $^{1}$CFCS, School of CS, Peking University \quad
  $^{2}$Galbot \quad
  $^{3}$Peking University \quad
  $^{4}$Renmin University of China
  \\[0.2em]
  \hspace{-1cm}
  $^{5}$The University of Hong Kong \quad
  $^{6}$Xiamen University Malaysia \quad
  $^{7}$Beihang University
}

\begin{document}
\maketitle
\setcounter{footnote}{0}

%===============================================================================

\begin{figure}[h!]
    \vspace{-2.0em}
    \centering
    \includegraphics[width=\textwidth]{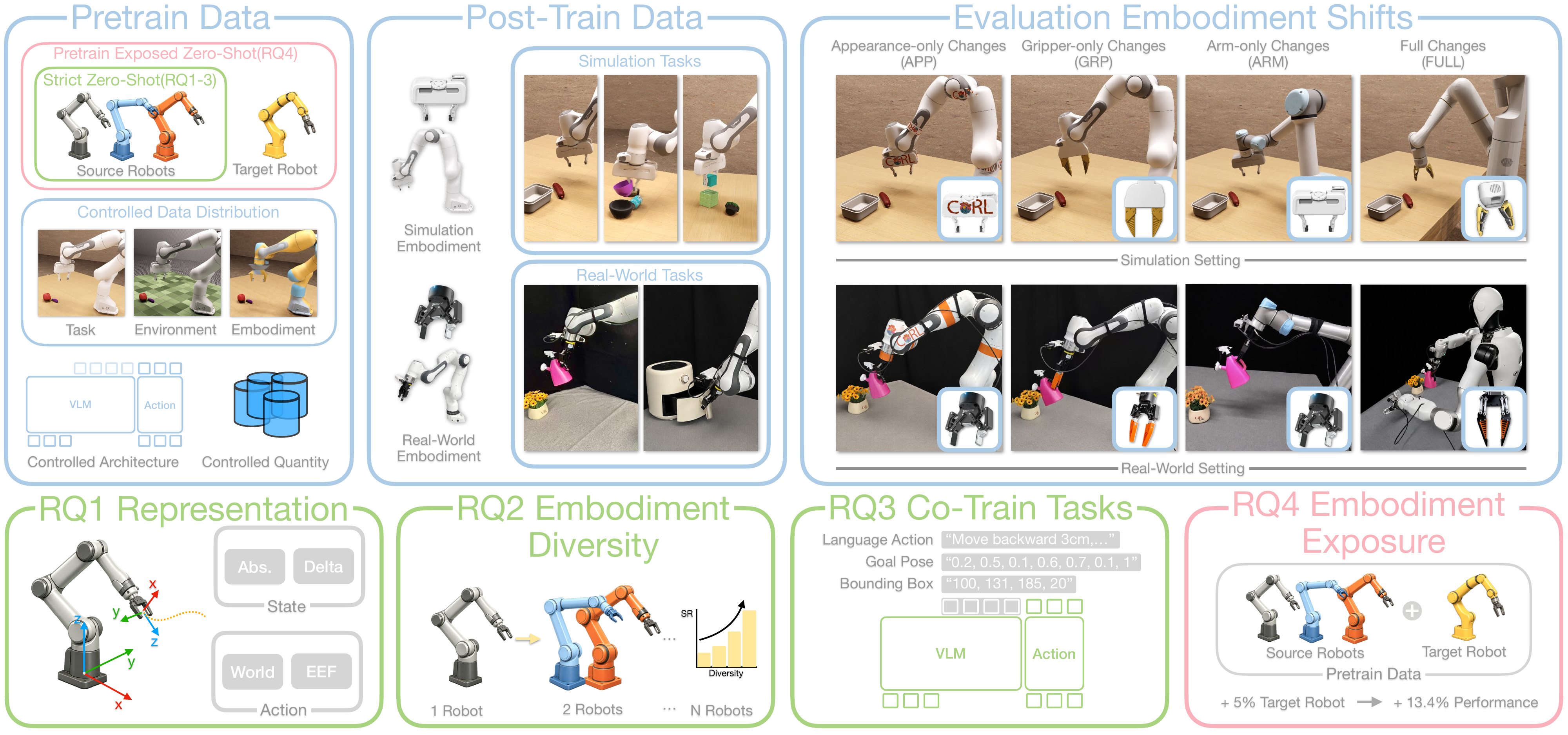}
    \vspace{-2.0em}
    \caption{Overview of our controlled study.
    We distinguish \emph{strict zero-shot transfer}, where the target robot is absent from all training data, from \emph{pretrain-exposed zero-shot transfer}.
    Under a controlled benchmark, we conduct fine-grained analyses across four embodiment shift categories, covering four research questions (RQ) on state-action representations, source-embodiment diversity, auxiliary co-training, and target-embodiment exposure.}
    \label{fig:teaser}
    \tightfloatafter
    \vspace{-0.5em}
\end{figure}

\begin{abstract}
Zero-shot generalization to unseen embodiments is important for generalizable vision-language-action (VLA) models as robot hardware evolves and task-specific data collection remains costly. 
However, a systematic understanding of this problem remains limited, in part because the literature lacks a unified zero-shot transfer definition and controlled evaluation settings that isolate embodiment changes from differences in tasks, scenes, or protocols. 
To address this gap, we first distinguish \emph{strict zero-shot transfer}, where the target embodiment is absent from all training data, from \emph{pretrain-exposed zero-shot transfer}, where it appears only during pretraining. We then introduce a controlled benchmark spanning 14 held-out target embodiments across simulation and real-world validation. 
Within this framework, we conduct a controlled analysis of four factors: state-action representations, pretraining embodiment diversity, auxiliary co-training objectives, and target-embodiment exposure. 
Experimental results show that local end-effector (EEF) state-action representations, the source embodiment diversity, and auxiliary co-training improve cross-embodiment transfer by around 15, 18, and 7 percentage points, respectively. 
We further find that adding only 5\% target-embodiment data during pretraining improves average target-embodiment progress by 13.4 percentage points, showing that strict and pretrain-exposed zero-shot transfer are distinct and should be reported separately. 
Together, these findings provide practical guidance for evaluating and improving cross-embodiment VLA transfer in stationary tabletop manipulation with two-finger grippers, while motivating future investigation of broader settings including mobile-base control, dexterous hands, and long-horizon tasks.
Our project page is at \href{https://theone2006.github.io/ZETA-Web/}{https://theone2006.github.io/ZETA-Web}.
\end{abstract}


%===============================================================================

\section{Introduction}
\vspace{-0.35cm}

Zero-shot cross-embodiment transfer is central to generalizable vision-language-action (VLA) models: policies trained on source robots are expected to generalize to new target robots without target-task demonstrations. 
As robot hardware continues to evolve, such transfer becomes increasingly important for avoiding costly task-specific data collection on every new platform.

Despite growing interest, a systematic understanding of this problem remains limited.
First, ``zero-shot transfer'' is used inconsistently to denote two protocols with different generalization difficulty: some keep the target embodiment absent from all training, while others include it in pretraining but exclude it from task-specific post-training.
Another challenge is that existing evaluations often couple embodiment shift together with differences in tasks, environments, cameras, data quantities, and protocols, making it difficult to attribute failures to embodiment mismatch itself.
In this work, we focus on stationary tabletop manipulation with two-finger grippers, allowing us to study embodiment changes in a controlled setting while holding the broader manipulation regime fixed.

To address the first challenge, we treat target inclusion during pretraining as an explicit experimental variable.
We define \emph{strict zero-shot transfer} as the setting where the target embodiment is absent from all training, and \emph{pretrain-exposed zero-shot transfer} as the setting where it appears only during pretraining.
These settings reflect two common deployment scenarios: newly designed robots with no collected data (e.g., a team iterates on its hardware and has data only from earlier robot versions), and publicly available robots with no private post-training data, respectively.

To address the second challenge, we construct a controlled simulation benchmark over seven held-out target embodiments within this tabletop manipulation setting, matching the backbone, downstream source data, and evaluation setting, with experiment-specific budget controls designed to characterize each factor.
We group embodiment shifts into appearance-only, gripper-only, arm-only, and full-embodiment changes to provide a more fine-grained analysis, with each simulation comparison evaluated on 6,300 rollouts per model.
We further complement this simulation study with real-world validation under matched categories.

Within this framework, our work studies four research questions (RQ) covering the major design axes of cross-embodiment transfer. 
RQ1: Which state-action representation best supports strict zero-shot transfer to unseen embodiments? 
RQ2: Under different pretraining budget controls, how does varying the number of source embodiments within a procedurally generated pool affect strict transfer?
RQ3: Do co-training objectives improve transfer beyond imitation learning alone? 
RQ4: How much does seeing the target embodiment during pretraining change the difficulty of zero-shot transfer? 
RQ1--RQ3 are evaluated under strict zero-shot transfer, while RQ4 compares the strict and pretrain-exposed settings.

The resulting simulation experiments show that all four factors substantially affect cross-embodiment transfer.
Local EEF-centered state-action representations improve strict transfer by around 15 percentage points on average.
Under a fixed 640K-trajectory pretraining budget, the 512-source setting within our procedurally generated Franka-style source pool outperforms the single-source model by around 18 percentage points; auxiliary co-training further raises average progress from 75.7\% to 82.3\%.
Finally, adding only 5\% target-embodiment data during pretraining improves average target-embodiment progress by 13.4 percentage points, showing that strict zero-shot transfer and pretrain-exposed zero-shot transfer should be reported separately.
For RQ1 and RQ2, real-world validation further shows the same overall trends.
% Finally, adding only 5\% target-embodiment data during pretraining improves full-embodiment success by 18.0 percentage points, showing that strict zero-shot transfer and pretrain-exposed zero-shot transfer should be reported separately.

In summary, we contribute: (i) \textbf{Protocol clarification}. We separate strict and pretrain-exposed zero-shot cross-embodiment transfer and argue that they should be reported separately; (ii) \textbf{Controlled benchmark}. We introduce a benchmark that isolates embodiment mismatch across four shift categories; (iii) \textbf{Factorized study}. We vary state-action representations, source-embodiment diversity, auxiliary co-training, and target-embodiment exposure to quantify their effects; and (iv) \textbf{Practical guidance}. We derive recommendations for evaluating and improving VLA transfer within stationary tabletop manipulation with two-finger grippers, favoring local EEF-frame representations, controlled source-diversity evaluation, and auxiliary co-training.

%===============================================================================

\tightsection{Related Work}

\tightsubsection{Data Composition in Multi-Embodiment Pretraining}
Robot foundation models increasingly rely on heterogeneous pretraining data.
Federated multi-embodiment corpora~\citep{openx2024rtx} pool demonstrations across institutions and robots, complemented by datasets spanning in-the-wild teleoperation~\citep{khazatsky2025droid}, skill-rich manipulation~\citep{fang2023rh20t,walke2024bridgedatav2}, multi-embodiment benchmarks~\citep{robomind}, large-scale real-world dual-arm platforms~\citep{agibotworld2025}, unified cross-robot collections~\citep{ario2024}, and high-fidelity synthetic data~\citep{interndataa1}.
These resources highlight the benefits of cross-robot data pooling, while generalist vision-language-action policies extend across heterogeneous platforms for adaptation and open-world manipulation~\citep{brohan2022rt1,zitkovich2023rt2,octo2024,kim2024openvla,black2024pi0,physicalintelligence2025pi05,nvidia2025gr00tn1,li2024cogact,wu2023gr1,li2023roboflamingo,shukor2025smolvla,deng2025graspvlagraspingfoundationmodel,xvla,spatialvla2025,kim2025openvlaoft}.
Other advances include plug-in action experts~\citep{dexvla}, diffusion-based policies~\citep{liu2026rdt2,liu2024rdt1b}, human-centric cross-embodiment learning~\citep{luo2026being}, multi-domain policies~\citep{crossformer}, multi-embodiment locomotion~\citep{urma,locoformer}, and latent-action learning from non-robot or weakly labeled videos~\citep{ye2025latentactionpretraining,univla}.
Studies of data diversity~\citep{diversityall2025} and embodiment scaling~\citep{embodimentscaling2025} further highlight the role of pretraining composition in transfer, but prior work often varies data scale, task distribution, embodiment coverage, and architecture simultaneously.
We instead isolate pretraining composition by examining procedural source-embodiment diversity under complementary fixed-total and fixed-per-embodiment budgets, and target-embodiment exposure under a fixed total pretraining budget.
\tightsubsection{Cross-Embodiment Policy Design: Representations, Interfaces, and Supervision}
Beyond data composition, transfer depends on how policies represent and map behavior across robots.
One line of work learns shared skill or action spaces from cross-embodiment demonstrations, human videos, image-space motion tracks, universal action codes, and task-centric latent actions~\citep{xskill,motif,uniskill,egovla,motiontracks,uniact,xlvla,univla,wang2024crossembodimentlatent,bauer2025latentactiondiffusion,mu2026opfa,yuan2024crossdex,yan2026unifiedlatent}.
Another adapts shared backbones through embodiment-specific interfaces or modules, including heterogeneous stems and action heads, soft prompts, unified control interfaces, latent guidance, shared hardware tools, and visual robot or viewpoint augmentation~\citep{hpt,xvla,capo2026,cei,ate,legato,mirage,roviaug,shadow2025,lepert2025phantom}.
Auxiliary supervision further provides transferable semantic, spatial, and motion-level structure through language-action prediction, co-training, chain-of-thought reasoning, and spatial-temporal grounding~\citep{lap,lin2026systematiccotraining,ecot,cotvla,tracevla}.
Recent work also examines the necessity of explicit proprioceptive state in visuomotor policies~\citep{zhao2025needproprioceptivestatesvisuomotor}.
While prior evaluations often vary datasets and protocols alongside policy design, we compare state-action representations and auxiliary co-training under matched architecture, data, and evaluation settings.
\tightsubsection{Zero-Shot Cross-Embodiment Evaluation and Benchmarks}
Prior benchmarks and studies examine cross-embodiment generalization, including manipulation and navigation transfer~\citep{yang2024pushing}, multi-embodiment manipulation data~\citep{robomind}, cross-embodiment manipulation benchmarks~\citep{anybody2025}, and the effects of data diversity~\citep{diversityall2025} and embodiment scaling~\citep{embodimentscaling2025}.
However, ``zero-shot'' may refer either to targets absent from all training or to targets seen during pretraining but not post-training.
Embodiment changes may also be confounded by task, scene, camera, data-budget, or protocol differences.
Our benchmark instead distinguishes strict and pretrain-exposed zero-shot transfer and isolates appearance-only, gripper-only, arm-only, and full-embodiment shifts under matched tasks, environments, cameras, data quantities, and training protocols.

%===============================================================================

\vspace{-0.35cm}
\section{Problem Formulation and Preliminaries}
\label{sec:problem}
\vspace{-0.35cm}

\textbf{Notation.}
Let $\mathcal{E}$ denote a set of embodiments and $\mathcal{T}$ denote a set of tasks.
An embodiment $e \in \mathcal{E}$ specifies robot-dependent factors such as appearance, morphology and kinematics.
For an embodiment--task pair $(e,t)$, let $\mathcal{D}(e,t)$ denote the corresponding demonstration dataset.
For sets $\mathcal{E}' \subseteq \mathcal{E}$ and $\mathcal{T}' \subseteq \mathcal{T}$, we write $ \mathcal{D}(\mathcal{E}', \mathcal{T}') = \bigcup_{(e,t)\in \mathcal{E}'\times \mathcal{T}'} \mathcal{D}(e,t)$, with singleton sets omitted when clear.
Subscripts $\mathrm{pre}$ and $\mathrm{post}$ denote pretraining and post-training, respectively.

\textbf{Transfer Setup.}
We study supervised imitation learning under a two-stage training protocol.
A policy is first pretrained on $\mathcal{D}_{\mathrm{pre}}$, whose embodiment and task composition depends on the research question.
Let $\mathcal{E}_{\mathrm{pre}}$ denote the embodiments appearing in $\mathcal{D}_{\mathrm{pre}}$.
It is then post-trained on downstream tasks using only a source embodiment $e_s$, i.e., $\mathcal{D}_{\mathrm{post}}=\mathcal{D}(e_s,\mathcal{T}_{\mathrm{post}})$.
Evaluation is performed on a target embodiment $e_t \neq e_s$ for the same downstream tasks $\mathcal{T}_{\mathrm{post}}$.

\textbf{Strict and Pretrain-Exposed Zero-Shot Protocols.}
We distinguish two zero-shot protocols according to target-embodiment exposure during pretraining.
In \emph{strict zero-shot transfer}, the target embodiment is absent from all training data: $e_t \notin \mathcal{E}_{\mathrm{pre}}$.
In \emph{pretrain-exposed zero-shot transfer}, $e_t \in \mathcal{E}_{\mathrm{pre}}$ but $e_t$ is absent from post-training.
Prior cross-embodiment studies use ``zero-shot'' to describe different exposure settings, complicating direct comparisons.
We formalize these distinctions to enable controlled evaluation and provide a taxonomy of prior work in Appendix~\ref{sec:zeroshot_taxonomy}.

\textbf{Embodiment Shift Taxonomy.}
To provide a more fine-grained analysis of cross-embodiment transfer, we group embodiment shifts into four categories.
\emph{Appearance-only} shifts change visual appearance while preserving morphology and kinematics.
\emph{Gripper-only} shifts alter end-effector geometry while keeping the arm fixed.
\emph{Arm-only} shifts change arm morphology or kinematics while preserving the end effector.
\emph{Full-embodiment} shifts change both arm/platform and end-effector factors.
Sec.~\ref{sec:experimental_design} instantiates these categories in simulation and real-world settings, with additional details provided in Appendix~\ref{sec:supp_embodiment_shift_taxonomy}.

%===============================================================================

\tightsection{Study Design}
\label{sec:experimental_design}

\tightsubsection{Representative State-Action Representations}
\label{sec:action_taxonomy}

Existing policies often use either joint-space or Cartesian EEF interfaces.
Because strict zero-shot transfer evaluates on target robots absent from all training data, joint-indexed representations may not share dimensionality, limits, or kinematic semantics across robots. We therefore focus on fixed-dimensional Cartesian end-effector representations.

\begin{wraptable}{r}{0.6\textwidth}
\vspace{-0.6cm}
\centering
\small
\setlength{\tabcolsep}{4.5pt}
\caption{State and action representations studied in RQ1.}
\label{tab:state_action_representation}
\resizebox{\linewidth}{!}{%
\begin{tabular}{ll}
\hline
\multicolumn{2}{l}{\textbf{Action representation}} \\
\hline
World-Delta action
& $\{\mathcal{V}(T_{b,e_{t'+1}})-\mathcal{V}(T_{b,e_{t'}})\}_{t'=t}^{t+K-1}$
\\
EEF-Delta action
& $\{\mathcal{V}(T_{b,e_{t'}}^{-1} \cdot T_{b,e_{t'+1}})=\mathcal{V}(T_{e_{t'},e_{t'+1}})\}_{t'=t}^{t+K-1}$ \\
\hline
\multicolumn{2}{l}{\textbf{State representation}} \\
\hline
Abs. EEF state
& $\{\mathcal{V}(T_{b,e_{t-i}})\}_{i=0}^{h}$
\\
EEF-Delta state
& $\{\mathcal{V}(T_{b,e_t}^{-1} \cdot T_{b,e_{t-i}})=\mathcal{V}(T_{e_t,e_{t-i}})\}_{i=0}^{h}$
\\
\hline
\end{tabular}%
}
\vspace{-0.5cm}
\end{wraptable}
As summarized in Table~\ref{tab:state_action_representation}, we compare two action frames and two state encodings.
We use $b$ to denote the robot base frame, $e_t$ to denote the gripper frame at time $t$, $T_{x,y}$ to denote the transform from frame $x$ to frame $y$, and $\mathcal{V}(T)$ to denote the vectorization of a transform $T$.
For $T=\begin{bmatrix} R & p \\[0.05em] 0 & 1 \end{bmatrix}$, we define $\mathcal{V}(T)=[p,\operatorname{Euler}(R)]$.
World-Delta and EEF-Delta actions differ in whether the next motion is expressed in the robot base frame or in the current gripper frame.
EEF-centered actions reduce dependence on embodiment-specific base definitions and better match the local motion semantics observed from a wrist camera.

A similar EEF-centered principle also applies to state representations.
Recent state-free policies~\cite{zhao2025needproprioceptivestatesvisuomotor} remove explicit end-effector pose history and report improved cross-embodiment transfer.
Under the notation in Table~\ref{tab:state_action_representation}, this design can be viewed as a degenerate EEF-Delta state: when $h=0$, the history contains only $T_{e_t,e_t}=I$, a constant identity transform that carries no state information, so state-free policies can be viewed as a degenerate EEF-Delta state.

\tightsubsection{Controlled Data Construction}
\begin{wrapfigure}{r}{0.23\textwidth}
    \vspace{0em}
    \centering
    \includegraphics[width=\linewidth]{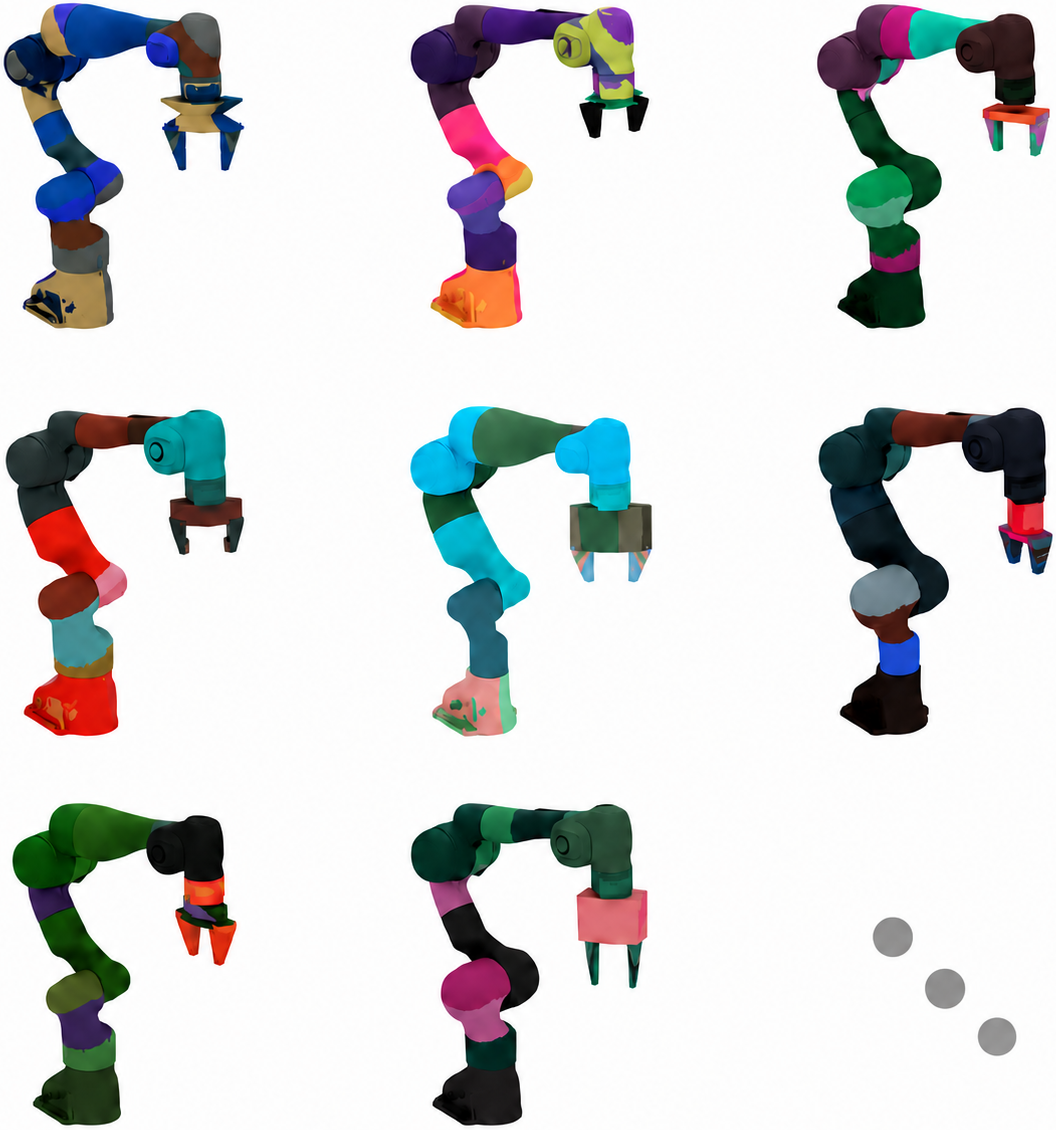}
    \caption{Procedurally generated embodiment pool for pretraining.}
    \label{fig:embodiments}
    \vspace{-1.2em}
\end{wrapfigure}
We clearly separate pretraining from post-training and impose $\mathcal{T}_{\mathrm{pre}} \cap \mathcal{T}_{\mathrm{post}}=\varnothing$, so downstream source-to-target ($e_s \to e_t$) embodiment transfer gap can be measured under controlled task exposure.
Unless otherwise specified, our primary comparisons use a fixed pretraining budget of 640K trajectories and post-train only on source-embodiment downstream data: 40K trajectories per task in simulation and 50 demonstrations per task in the real world.
RQ2 additionally includes a complementary fixed-per-embodiment-budget sweep, in which the total pretraining budget varies with the number of source embodiments.

To control non-embodiment factors, we use a simulator to construct a large-scale pretraining dataset with controlled data distributions.
Following~\cite{locoformer, embodimentscaling2025}, we construct 512 procedurally generated Franka-family source embodiments by introducing geometric and visual variation.
Fig.~\ref{fig:embodiments} shows representative samples from this procedurally generated source-embodiment pool.
The seven commercially used test robots are imported as external held-out targets and excluded from this source pool, ensuring that strict zero-shot evaluations use target embodiments absent from all training data.
Inspired by prior work~\cite{deng2025graspvlagraspingfoundationmodel, interndataa1, yin2026geniesim30, maddukuri2025simandrealcotrainingsimplerecipe} that perform direct sim-to-real transfer, we apply domain randomization to visual properties (camera poses, lighting, background, and layout) and physical properties (mass and friction) under simple quasi-static assumptions, reducing the impact of sim-to-real gaps on our experimental analysis.
We further complement the simulation evaluation with real-world validation using real post-training data and real-world trials. Details are provided in Appendix~\ref{sec:supp_data_construction}.

\tightsubsection{Auxiliary Co-training Objectives}
\label{sec:cotrain_objectives}

Motivated by prior VLA work showing that auxiliary supervision adds transferable semantic supervision beyond imitation learning~\cite{zitkovich2023rt2, liu2024robomambaefficientvisionlanguageactionmodel, physicalintelligence2025pi05, lin2026onetwovlaunifiedvisionlanguageactionmodel, qu2026eo1openunifiedembodied, chen2025internvlam1spatiallyguidedvisionlanguageaction, lap, ecot, lin2026systematiccotraining, cotvla, tracevla}, we include representative tasks as controlled study variants rather than default components of all models.
LAP predicts structured language-action descriptions, exposing policies to low-level motion semantics in the VLM text space.
Subgoal prediction supervises task progress with a next-step goal representation, emphasizing transferable spatial and state-change structure.
Task-conditioned bounding-box prediction grounds instructions in object-centric image regions, testing whether explicit spatial grounding improves transfer.
When enabled, auxiliary objectives are applied during both pretraining and post-training.

\begin{figure}[htbp]
    \tightfloatbefore
    \centering
    \includegraphics[width=0.62\linewidth]{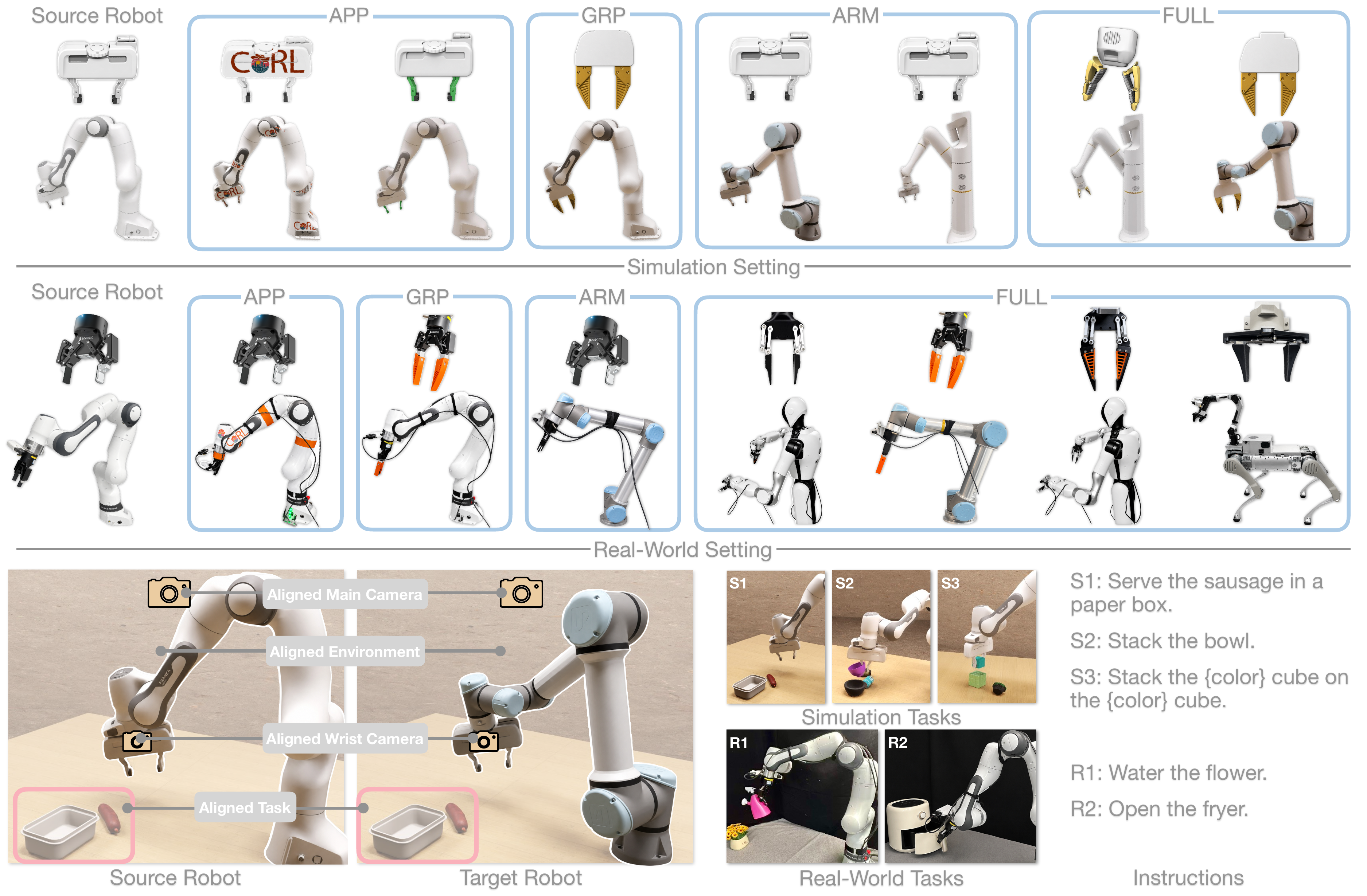}
    \caption{Evaluation setup for controlled cross-embodiment transfer. We evaluate seven held-out target embodiments across four shift categories in simulation and the real world, carefully matching post-training and test scenes to align non-embodiment factors (cameras, environments, and tasks).}
    \label{fig:experimental_setup}
    \vspace{-0.3em}
\end{figure}

\tightsubsection{Embodiment Shift Categories}
We focus on common two-finger-gripper embodiments because they cover substantial appearance, morphology, and kinematic variation while keeping the end-effector interface comparable for in-depth analysis.
We instantiate the shift taxonomy from Sec.~\ref{sec:problem}; additional embodiment details are provided in Appendix~\ref{sec:supp_embodiment_shift_taxonomy}.
Fig.~\ref{fig:experimental_setup} summarizes the held-out target embodiments and their grouping into these four shift categories.
In simulation, we use Franka as the source robot; appearance-only shifts modify only the Franka texture (Franka with logo and Franka with green fingers).
These target appearances are near the source visual distribution but remain held out from pretraining (Appendix~\ref{sec:supp_data_construction}).
Gripper-only shifts replace the end-effector geometry while keeping the Franka arm (Franka with UMI gripper).
Arm-only shifts replace the arm while keeping a Franka-style hand (UR5e and Google Robot with a Franka hand), and full-embodiment shifts replace all configuration (UR5e+UMI and GoogleRobot).
The real-world setup mirrors these categories using Franka-Robotiq as the source embodiment and matched hardware changes for each shift type.

%===============================================================================

\tightsection{Controlled Experiments}
\label{sec:experiments}

We organize the experiments around four controlled comparisons.
For each research question, we first define the variable under study and the fixed conditions, then report the corresponding transfer results across embodiment-shift categories.

\tightsubsection{Shared Experimental Setup}

\textbf{Evaluation protocol.}
To isolate embodiment shift, we align non-embodiment factors between evaluation and post-training as closely as possible, and we keep the initial evaluation scene consistent across all models.
For each downstream task, the test objects, initial-pose sampling region ($30\,\mathrm{cm} \times 40\,\mathrm{cm}$), background, table layout, and lighting are kept consistent with the corresponding post-training setting.
Both simulation and real-world evaluation use two visual observations: one wrist-mounted view and one third-person view.
We further align these camera poses so that, under a canonical robot pose, the image-plane gripper distribution remains close to the post-training distribution.
Additional calibration and setup details are provided in Appendix~\ref{sec:supp_experimental_setup_details}.

\textbf{Model backbone and training.}
We instantiate $\pi_\theta$ with a representative VLA backbone following $\pi_{0.5}$~\cite{physicalintelligence2025pi05}.
Each model follows the same two-stage protocol: pretraining on $\mathcal{D}_{\mathrm{pre}}$ followed by post-training on $\mathcal{D}_{\mathrm{post}}$.
All comparisons keep the backbone, input configuration, action horizon, losses, data budget, and optimization schedule fixed unless explicitly stated; model training details are provided in Appendix~\ref{sec:supp_model_training_details}.

\begin{table*}[t]
\tightfloatbefore
\centering
\small
\setlength{\tabcolsep}{6pt}
\caption{RQ1 zero-shot simulation results: EEF-Delta states improve arm-only and full-embodiment transfer, while EEF-Delta actions yield better average score.}
\label{tab:rq1}
\resizebox{1.0\textwidth}{!}{%
\begin{tabular}{lllccccc}
\hline
\rule{0pt}{2.6ex}\textbf{State}
& \rule{0pt}{2.6ex}\textbf{Action}
& \rule{0pt}{2.6ex}\textbf{$\mathbf{e_s}$}
& \rule{0pt}{2.6ex}\textbf{APP}
& \rule{0pt}{2.6ex}\textbf{GRP}
& \rule{0pt}{2.6ex}\textbf{ARM}
& \rule{0pt}{2.6ex}\textbf{FULL}
& \rule{0pt}{2.6ex}\textbf{Average} \\
\hline
% Qpos         & Delta qpos      & -- & -- & -- & -- & -- & -- \\
Abs. EEF & World-Delta  & 87.7 $\pm$ 2.1 & 86.3 $\pm$ 1.3 & 82.6 $\pm$ 1.7 & 38.5 $\pm$ 0.8 & 33.9 $\pm$ 2.3 & 60.3 $\pm$ 1.1 \\
Abs. EEF & EEF-Delta    & \textbf{92.6} $\pm$ 1.0 & \textbf{91.8} $\pm$ 1.1 & \textbf{88.2} $\pm$ 2.6 & 39.9 $\pm$ 0.5 & 38.6 $\pm$ 0.8 & 64.6 $\pm$ 0.4\\
EEF-Delta & World-Delta  & 87.8 $\pm$ 1.4 & 88.0 $\pm$ 0.9 & 77.2 $\pm$ 1.0 & 69.8 $\pm$ 1.9 & 58.4 $\pm$ 1.3 & 73.4 $\pm$ 0.7 \\
EEF-Delta & EEF-Delta    & 91.5 $\pm$ 0.6 & 90.0 $\pm$ 1.1 & 78.3 $\pm$ 2.4 & \textbf{74.3} $\pm$ 0.9 & \textbf{60.4} $\pm$ 0.9 & \textbf{75.7} $\pm$ 1.3 \\
\hline
\end{tabular}%
}
\tightfloatafter
\end{table*}

\textbf{Metrics and evaluation scale.}
We score each rollout by task progress in $[0,100]$, with $100\%$ denoting full completion and partial credit assigned to necessary subgoals, such as contacting the air-fryer handle before opening.
All tables and plots report mean task progress as a percentage.
Simulation is our primary evaluation platform because it supports controlled, high-throughput comparisons across tasks, embodiments, and repeated trials.
For each model, we evaluate 3 downstream tasks on 7 test embodiments, with 100 rollouts per task--embodiment pair over 3 independent training runs, yielding $3 \times 7 \times 100 \times 3 = 6300$ simulation rollouts per model; we report mean task progress $\pm$ standard deviation across runs.
Each shift-category score aggregates all embodiments in that category and all tasks.
Real-world evaluation serves as an external-validity check: because physical trials require hardware setup, resets, and safety checks, we evaluate 2 downstream tasks on 7 test embodiments with 10 trials per task--embodiment pair, yielding $2 \times 7 \times 10 = 140$ real-world rollouts per model.

\tightsubsection{RQ1: How Do State-Action Representations Affect Transfer?}
RQ1 evaluates how the state-action representation defined in Sec.~\ref{sec:action_taxonomy} affects strict zero-shot transfer.
Actions can be expressed either as \textbf{World-Delta} or \textbf{EEF-Delta}, while state can be expressed either as \textbf{Abs. EEF} or as \textbf{EEF-Delta}.
Their pairwise combinations yield four state-action representations.
We evaluate these four variants using the pretraining data of the full set of 512 source embodiments and the same downstream post-training data.
While several embodied foundation models use mixed-representation to maximize the utilization of all datasets, we exclude it from this controlled comparison to isolate the effect of each state-action design choice.

\begin{wraptable}{r}{0.65\textwidth}
\vspace{-0.6cm}
\centering
\small
\setlength{\tabcolsep}{6pt}
\caption{Real-world results for RQ1. Each entry averages task progress on \emph{open the fryer} and \emph{water the flower}.}
\label{tab:rq1_real}
\resizebox{\linewidth}{!}{%
\begin{tabular}{lllccccc}
\hline
\rule{0pt}{2.6ex}\textbf{State}
& \rule{0pt}{2.6ex}\textbf{Action}
& \rule{0pt}{2.6ex}\textbf{$\mathbf{e_s}$}
& \rule{0pt}{2.6ex}\textbf{APP}
& \rule{0pt}{2.6ex}\textbf{GRP}
& \rule{0pt}{2.6ex}\textbf{ARM}
& \rule{0pt}{2.6ex}\textbf{FULL}
& \rule{0pt}{2.6ex}\textbf{Average} \\
\hline
Abs. EEF & World-Delta  & 87.5 & 87.5 & 67.5 & 47.5 & 21.3 & 56.0 \\
Abs. EEF & EEF-Delta    & 90.0 & 90.0 & 80.0 & 60.0 & 16.3 & 61.6 \\
EEF-Delta & World-Delta  & 85.0 & 77.5 & 60.0 & 62.5 & 43.1 & 60.8 \\
EEF-Delta & EEF-Delta    & \textbf{100.0} & \textbf{97.5} & \textbf{92.5} & \textbf{87.5} & \textbf{81.9} & \textbf{89.9} \\
\hline
\end{tabular}%
}
\vspace{-0.4cm}
\end{wraptable}

Table~\ref{tab:rq1} shows that EEF representations improve strict zero-shot transfer.
Switching action from world to eef raises the progress score from 60.3\% to 64.6\% with absolute state.
Replacing Abs. EEF state with EEF-Delta raises arm-only transfer from 38.5--39.9\% to 69.8--74.3\% and full-embodiment transfer from 33.9--38.6\% to 58.4--60.4\%, suggesting that local state preserve more transferable motion structure.
Real-world validation in Table~\ref{tab:rq1_real} follows the same pattern: EEF-Delta state with EEF-Delta actions achieves the best average score, improving from 56.0--61.6\% for the other representations to 89.9\%.
We therefore use this representation for RQ2 and RQ3.

\begin{figure}[t]
\tightfloatbefore
\centering
\includegraphics[width=\linewidth]{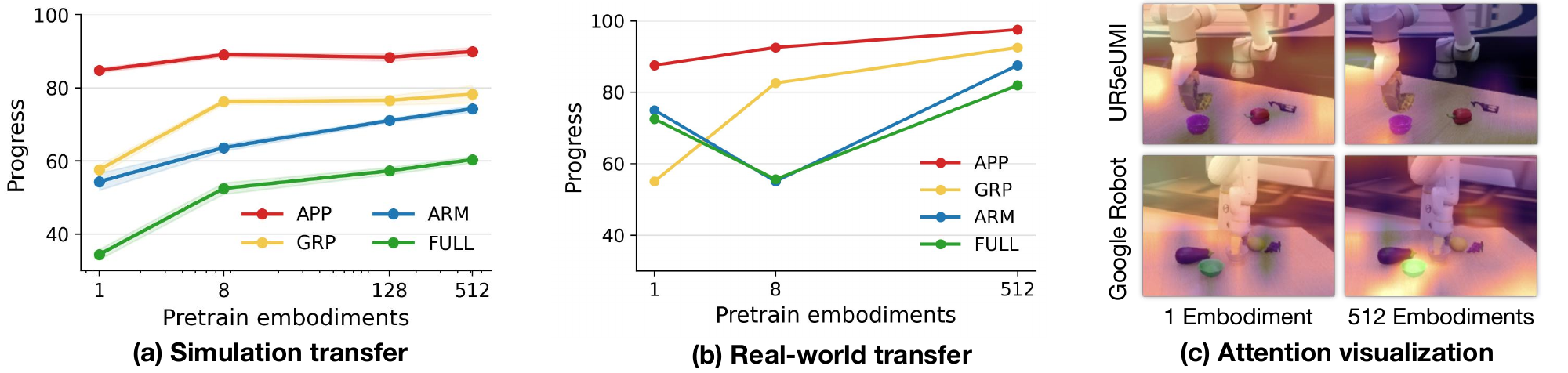}
\caption{RQ2: effect of source-embodiment diversity under a fixed pretraining budget.
The 512-source setting improves over the single-source setting in both simulation (a) and the real world (b).
Embodiment-diverse pretraining also shifts attention from diffuse robot--scene regions toward task-relevant objects and the gripper.}
% \caption{RQ2: effect of source-embodiment diversity under a fixed pretraining budget.
% Scaling source embodiments improves transfer in both simulation (a) and the real world (b), while shifting attention from diffuse robot--scene regions toward task-relevant objects and the gripper.}
\label{fig:rq2}
\tightfloatafter
\end{figure}

% \begin{figure}[htbp]
% \tightfloatbefore
% \centering
% \vspace{1.0em}
% \begin{minipage}[t]{0.4\linewidth}
%   \centering
%   \includegraphics[width=0.8\linewidth]{images/rq2/fix_per_emb_budget_sim.png}\\[-0.2em]
%   \small (a) Simulation
% \end{minipage}
% \hfill
% \begin{minipage}[t]{0.4\linewidth}
%   \centering
%   \includegraphics[width=0.8\linewidth]{images/rq2/fix_per_emb_budget_real.png}\\[-0.2em]
%   \small (b) Real world
% \end{minipage}
% \caption{RQ2 under a fixed per-embodiment pretraining budget.
% We sweep the number of source embodiments among 32, 128, and 512 while keeping the per-embodiment trajectory count fixed, so the total pretraining budget grows with $|\mathcal{E}_{\mathrm{pre}}|$.
% Average zero-shot progress is reported in simulation (a) and real-world evaluation (b).
% }
% \label{fig:rq2_fixed_per_emb_budget}
% \tightfloatafter
% \vspace{0.2em}
% \end{figure}
\begin{wrapfigure}{r}{0.3\linewidth}
  \centering
  \vspace{-1.5em}
  \includegraphics[width=\linewidth]{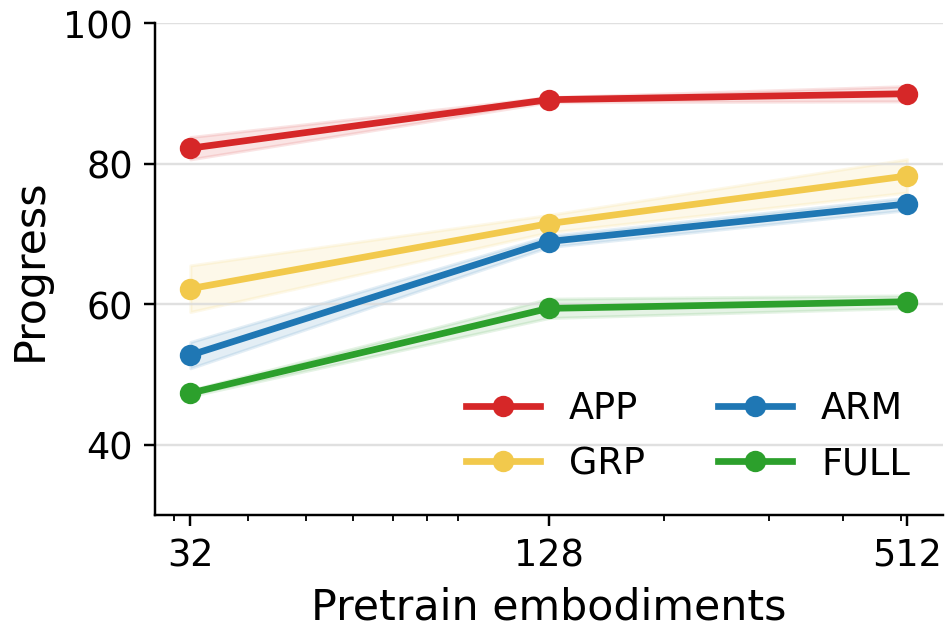}\\[-0.4em]
  {\footnotesize (a) Simulation}\\[0.2em]
  \includegraphics[width=\linewidth]{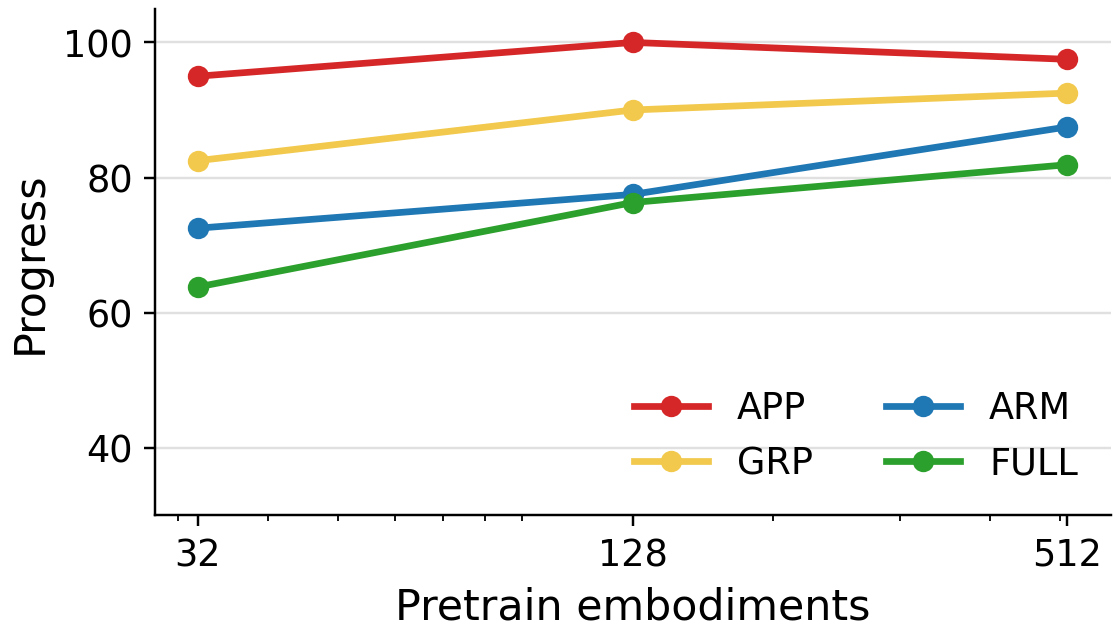}\\[-0.4em]
  {\footnotesize (b) Real world}
  \caption{RQ2 under a fixed per-embodiment pretraining budget.
  We sweep the number of source embodiments among 32, 128, and 512 while keeping the per-embodiment trajectory count fixed, so the total pretraining budget grows with $|\mathcal{E}_{\mathrm{pre}}|$.
  Average zero-shot progress is reported in simulation (a) and real-world evaluation (b).}
  \label{fig:rq2_fixed_per_emb_budget}
  \vspace{-1.8em}
\end{wrapfigure}

\tightsubsection{RQ2: How Does Source-Embodiment Diversity Affect Transfer Under Different Budget Controls?}

RQ2 studies procedural source-embodiment diversity under two complementary budget controls.
Our source pool consists of procedurally generated Franka-style embodiments, as described in Appendix~\ref{sec:supp_data_construction}.
In the primary comparison, we keep the total pretraining budget fixed at 640K pick-and-place trajectories while varying the number of source embodiments.
Consequently, the per-embodiment data decreases from 640K trajectories with a single source embodiment to 1,250 trajectories per embodiment with all 512 sources.
To complement this fixed-total-budget comparison, we conduct a second sweep that fixes the per-embodiment trajectory budget while varying the source pool over 32, 128, and 512 embodiments.
In this setting, each source embodiment contributes the same amount of pretraining data, so the total pretraining budget grows with $|\mathcal{E}_{\mathrm{pre}}|$ while all other factors remain unchanged.
All RQ2 models use the best-average state-action representation from RQ1, with downstream post-training and evaluation held fixed.

Fig.~\ref{fig:rq2}(a) shows that the 512-source model outperforms the single-source model, with the largest gains on gripper-only, arm-only, and full-embodiment shifts.
Attention visualizations further show that the 512-source model concentrates more strongly on task-relevant objects and the gripper than the single-source model, suggesting stronger task-relevant visual grounding.
A complementary UMAP visualization suggests greater cross-embodiment feature overlap with 512 source embodiments (Appendix~\ref{sec:rq2_umap}, Fig.~\ref{fig:rq2_feature_umap}).
In real-world validation, Fig.~\ref{fig:rq2}(b) reveals a non-monotonic intermediate regime: with 8 sources, arm-only and full-embodiment transfer falls below the single-source setting before recovering at 512 sources.
This pattern may reflect a fixed-budget trade-off in which increasing source coverage reduces per-embodiment data density before the source pool becomes sufficiently broad to support robust cross-embodiment transfer.
The complementary fixed-per-embodiment sweep in Fig.~\ref{fig:rq2_fixed_per_emb_budget} shows increasing average progress as the source pool grows from 32 to 128 to 512 embodiments in simulation, with the same overall trend in real-world evaluation.

Together, the two budget-control settings show that broader source coverage within our procedurally generated Franka-style embodiment pool can improve transfer, while the measured effect depends on how the pretraining budget is allocated.
Under a fixed total budget, increasing the source count trades off against per-embodiment data density and produces a non-monotonic real-world trend; under a fixed per-embodiment budget, performance improves over the evaluated 32--512 source range.
These results therefore support treating source-embodiment diversity and data-budget allocation as coupled design variables rather than assuming a universal monotonic benefit from increasing the number of source embodiments.

\tightsubsection{RQ3: How Do Co-training Tasks Affect Transfer?}

RQ3 evaluates whether the auxiliary co-training objectives defined in Sec.~\ref{sec:cotrain_objectives} further improve strict zero-shot transfer after using the best-average state-action representation from RQ1 and the full 512-embodiment pretraining pool from RQ2.
We evaluate these variants in simulation because several objectives require annotations beyond standard demonstrations.

Table~\ref{tab:rq3} shows that all auxiliary objectives improve over the no-co-training baseline, raising average progress from 75.7\% to 77.7--82.3\%, with gains concentrated on gripper-only, arm-only, and full-embodiment shifts.
Among language-action variants, the EEF-only objective outperforms the mixed-frame LAP variant, suggesting that auxiliary supervision is most effective when its action semantics align with the local EEF representation favored by RQ1.
\begin{table*}[htbp]
\tightfloatbefore
\centering
\small
\setlength{\tabcolsep}{6pt}
\caption{RQ3 simulation results: auxiliary co-training objectives. All co-training variants improve over imitation-only training, with the largest gains under larger embodiment shifts.}
\label{tab:rq3}
\resizebox{0.99\textwidth}{!}{%
\begin{tabular}{llccccc}
\hline
\rule{0pt}{2.6ex}\textbf{Post train data}
& \rule{0pt}{2.6ex}\textbf{$e_s$}
& \rule{0pt}{2.6ex}\textbf{APP}
& \rule{0pt}{2.6ex}\textbf{GRP}
& \rule{0pt}{2.6ex}\textbf{ARM}
& \rule{0pt}{2.6ex}\textbf{FULL}
& \rule{0pt}{2.6ex}\textbf{Average} \\
\hline
No co-training & \textbf{91.5} $\pm$ 0.6 & 90.0 $\pm$ 1.1 & 78.3 $\pm$ 2.4 & 74.3 $\pm$ 0.9 & 60.4 $\pm$ 0.9 & 75.7 $\pm$ 1.3 \\
% No co-training & \textbf{91.5} $\pm$ 0.6 & 90.0 $\pm$ 1.1 & 79.1 $\pm$ 1.4 & 74.3 $\pm$ 0.9 & 60.4 $\pm$ 0.9 & 75.9 $\pm$ 1.0 \\
Language-action (eef-frame) & 90.5 $\pm$ 1.1 & 88.6 $\pm$ 0.2 & 86.7 $\pm$ 1.4 & 79.3 $\pm$ 0.9 & 70.5 $\pm$ 0.5 & 81.3 $\pm$ 0.5 \\
% Language-action (eef-frame) & 90.5 $\pm$ 1.1 & 88.6 $\pm$ 0.2 & \textbf{87.1} $\pm$ 1.2 & 79.3 $\pm$ 0.9 & 70.5 $\pm$ 0.5 & 81.4 $\pm$ 0.7 \\
Language-action (both frames) & 89.5 $\pm$ 1.5 & 88.6 $\pm$ 0.4 & 79.8 $\pm$ 0.8 & 73.8 $\pm$ 1.6 & 68.4 $\pm$ 1.5 & 77.7 $\pm$ 0.3 \\
% Language-action (both frames) & 89.5 $\pm$ 1.5 & 88.6 $\pm$ 0.4 & 81.9 $\pm$ 0.8 & 73.8 $\pm$ 1.6 & 68.4 $\pm$ 1.5 & 78.2 $\pm$ 0.2 \\
Subgoal & 89.1 $\pm$ 1.8 & \textbf{90.3} $\pm$ 0.8 & \textbf{87.4} $\pm$ 2.2 & 77.8 $\pm$ 0.5 & 71.5 $\pm$ 1.5 & 81.8 $\pm$ 0.8 \\
% Subgoal & 89.1 $\pm$ 1.8 & \textbf{90.3} $\pm$ 0.8 & 85.7 $\pm$ 1.7 & 77.8 $\pm$ 0.5 & 71.5 $\pm$ 1.5 & 81.3 $\pm$ 0.7 \\
Task-conditioned BBox & 90.4 $\pm$ 1.4 & 90.2 $\pm$ 0.4 & 87.0 $\pm$ 0.9 & \textbf{80.1} $\pm$ 0.8 & \textbf{71.9} $\pm$ 1.4 & \textbf{82.3} $\pm$ 0.2 \\
% Task-conditioned BBox & 90.4 $\pm$ 1.4 & 90.2 $\pm$ 0.4 & 85.5 $\pm$ 1.0 & \textbf{80.1} $\pm$ 0.8 & \textbf{71.9} $\pm$ 1.4 & \textbf{81.9} $\pm$ 0.1 \\
\hline
\end{tabular}%
}
\tightfloatafter
\end{table*}

\tightsubsection{RQ4: How Does Target-Embodiment Exposure Affect Transfer?}

RQ4 studies \textit{pretrain-exposed zero-shot transfer} on two representative targets, UR5eUMI and GoogleRobot.
For each target-exposure ratio, we keep the total pretraining budget fixed at 640K trajectories by replacing a controlled fraction of source data with target-embodiment pretraining data; target downstream tasks remain excluded from all training.
We also report a non-zero-shot oracle initialized from the shared backbone pretrained on the full pretraining dataset
$\mathcal{D}(\mathcal{E}_{\mathrm{pre}}, \mathcal{T}_{\mathrm{pre}})$
and subsequently post-trained on the target-embodiment downstream data
$\mathcal{D}(e_t, \mathcal{T}_{\mathrm{post}})$, serving as an upper reference.

\begin{figure}[htbp]
\tightfloatbefore
\centering
\begin{minipage}[t]{0.47\linewidth}
  \centering
  \includegraphics[width=\linewidth]{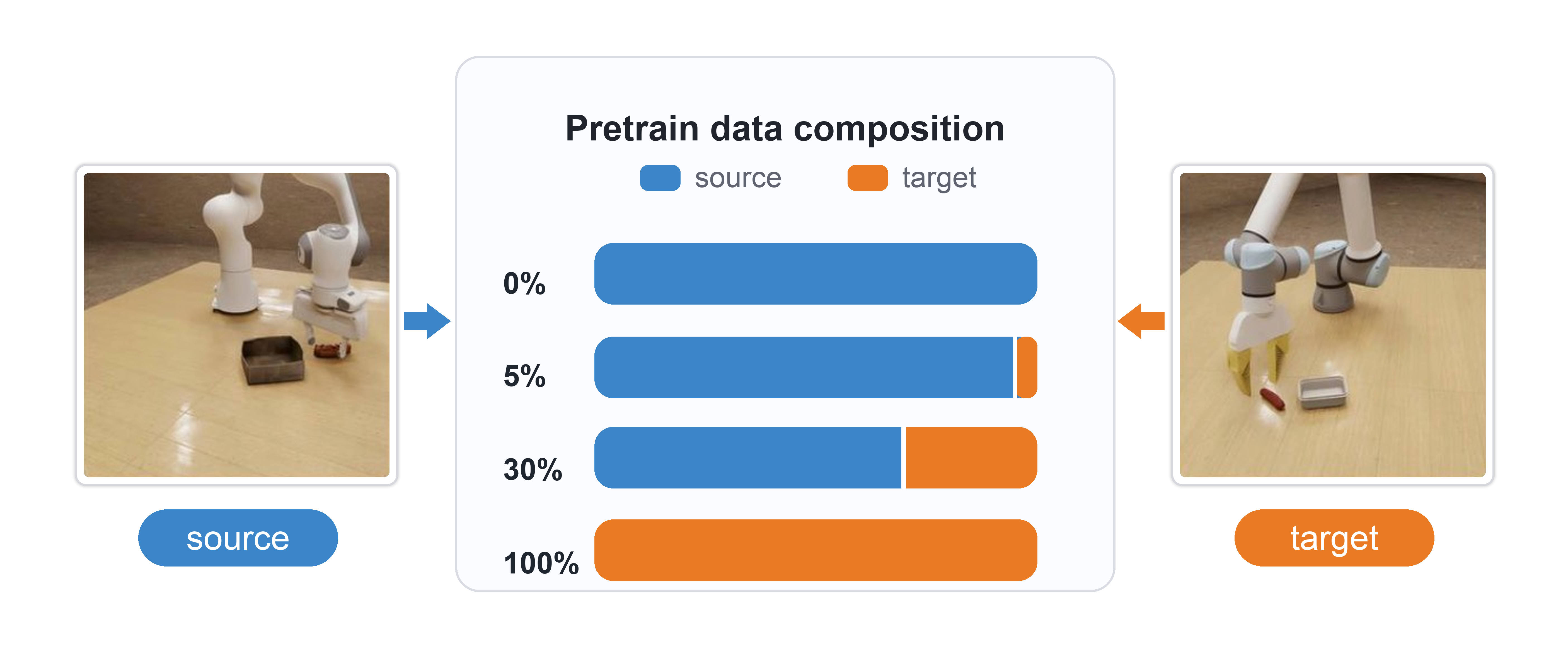}\\[-0.2em]
  \small (a) RQ4 pretraining composition setting
\end{minipage}
\hfill
\begin{minipage}[t]{0.24\linewidth}
  \centering
  \includegraphics[width=\linewidth]{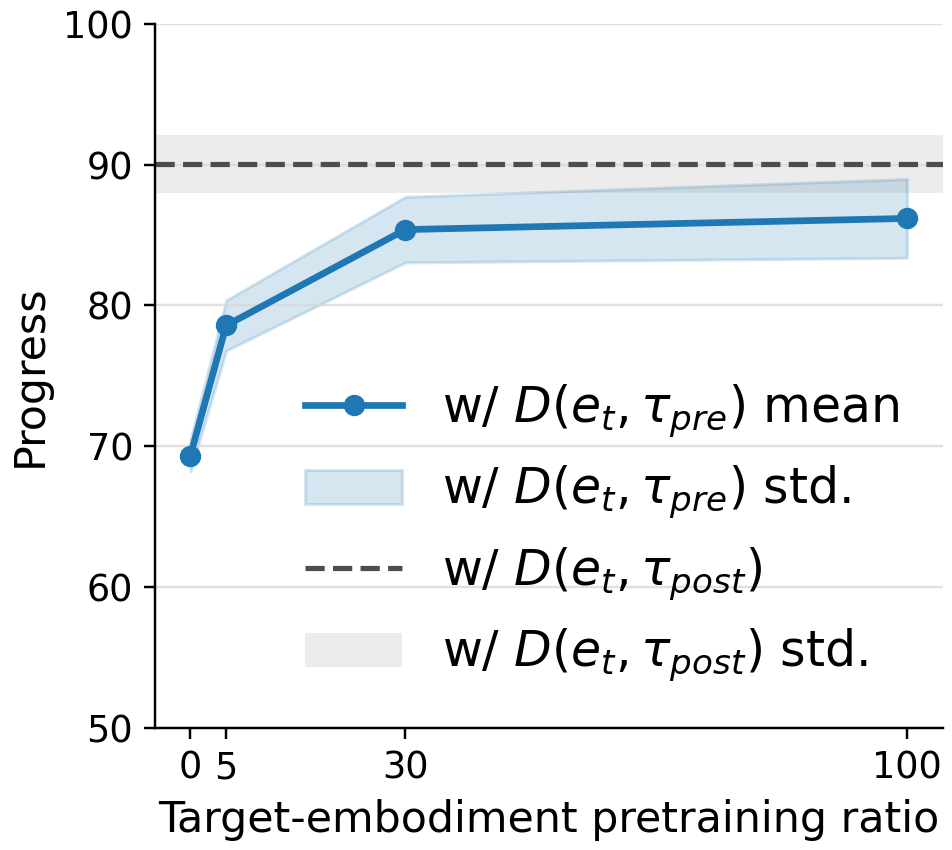}\\[-0.2em]
  \small (b) UR5eUMI
\end{minipage}
\hfill
\begin{minipage}[t]{0.24\linewidth}
  \centering
  \includegraphics[width=\linewidth]{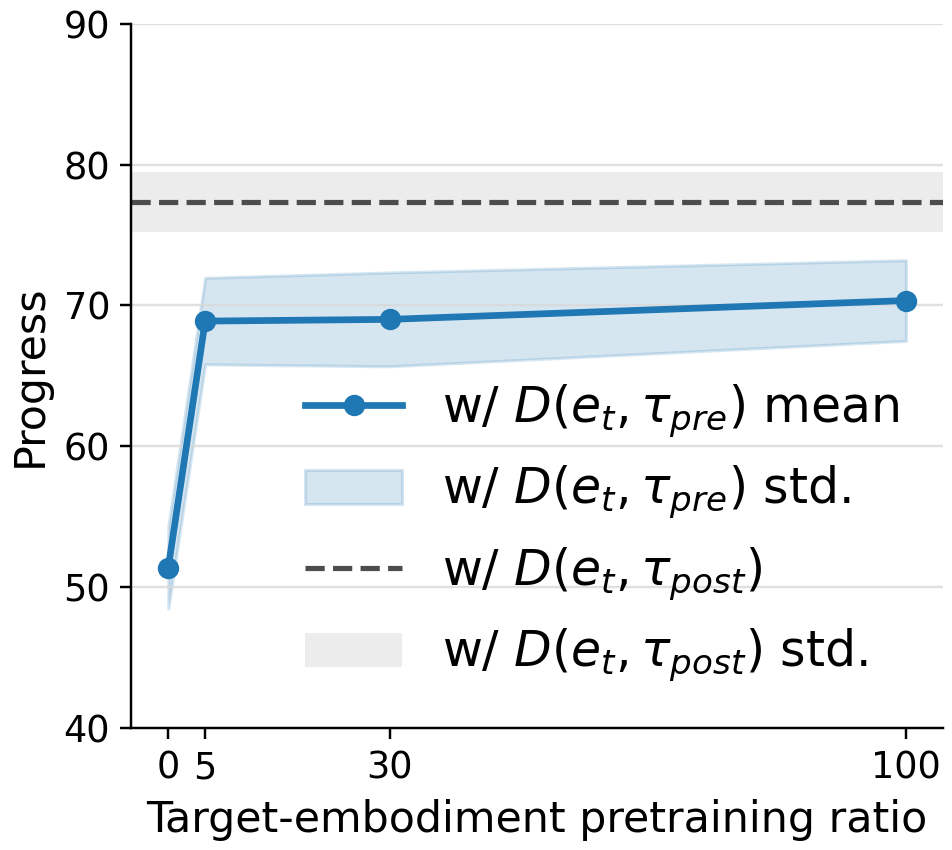}\\[-0.2em]
  \small (c) Google Robot
\end{minipage}
\vspace{-0.1em}
\caption{RQ4: target-embodiment exposure during pretraining. Only 5\% of target embodiment pretraining data substantially reduce the transfer gap on both UR5eUMI and Google Robot, but remain below the oracle performance of post-training on the target embodiment and test tasks.}
\label{fig:rq4_plots}
\tightfloatafter
\vspace{0.3em}
\end{figure}

Fig.~\ref{fig:rq4_plots} shows that even limited target embodiment exposure reduces the transfer gap.
Adding only 5\% target-embodiment data during pretraining yields the largest marginal gain, improving UR5eUMI from 69.3\% to 78.6\% and GoogleRobot from 51.4\% to 68.9\%.
However, both curves remain below the oracle, indicating that target-embodiment pretraining mitigates but does not eliminate the need for target-task adaptation.
These results support reporting strict and pretrain-exposed zero-shot transfer separately, since target inclusion during pretraining changes the difficulty of the evaluation even when downstream target tasks are excluded.

\begin{wraptable}{r}{0.42\columnwidth}
\vspace{-0.35cm}
\centering
\footnotesize
\setlength{\tabcolsep}{4pt}
\caption{Interaction between state-action representation and target-embodiment exposure. Average progress (\%) on the RQ4 targets.}
\label{tab:rq4_target_exposure}
\begin{tabular}{llcc}
\toprule
\textbf{State} & \textbf{Action} & \textbf{0\%} & \textbf{5\%} \\
\midrule
Abs. EEF & World-Delta & 33.9 & 66.9 \\
Abs. EEF & EEF-Delta & 38.6 & 69.2 \\
EEF-Delta & World-Delta & 58.4 & 67.5 \\
EEF-Delta & EEF-Delta & \textbf{60.4} & \textbf{71.1} \\
\bottomrule
\end{tabular}
\vspace{-0.3cm}
\end{wraptable}
We further test whether the effect of target-embodiment exposure depends on the state-action representation studied in RQ1.
Table~\ref{tab:rq4_target_exposure} compares all four representations under strict zero-shot transfer and 5\% target-embodiment exposure.
Target exposure improves all four representations, while substantially narrowing their performance gaps.
This suggests that representation choice matters most under strict transfer, whereas even limited target-embodiment exposure can partially compensate for less transferable representations.

\tightsubsection{Practical Guidance for Cross-Embodiment Transfer within the Studied Setting}
Taken together, these research questions suggest a practical recipe for evaluating and improving cross-embodiment VLA transfer.
First, zero-shot claims should specify whether the target embodiment is absent from all training data or only absent from task-specific post-training, because these conditions can differ substantially in transfer difficulty.
Second, for transfers with large embodiment gaps, we recommend embodiment-compatible representations that use both EEF-Delta state and actions, while absolute EEF state with EEF-Delta actions can remain competitive when the shift is limited to the end effector or appearance.
Third, source-embodiment diversity should be considered jointly with data-budget allocation. Within our procedurally generated Franka-style source pool, the 512-source setting yields the strongest overall transfer.
Finally, auxiliary co-training objectives generally improve transfer.

%===============================================================================

\tightsection{Limitations and Conclusion}
\label{sec:limitations_conclusion}

\textbf{Limitations.}
We study cross-embodiment transfer in a controlled setting of stationary tabletop manipulation with representative two-finger-gripper manipulators.
This enables matched comparisons that isolate embodiment differences but excludes broader robot morphologies and manipulation settings.
Extending to mobile manipulation with base motion and dexterous-hand manipulation remains an important direction, particularly given the latter's more complex contact topology, action dimensionality, and skill structure.
Our evaluation also focuses on relatively short-horizon tasks, leaving open whether the observed transfer patterns persist over longer horizons.
Finally, our experiments characterize policy behavior and cross-embodiment transfer rather than optimize all system-level components required for deployment.
These choices enable focused, comparable analysis while limiting the scope of our conclusions.

\textbf{Conclusion.}
We present a controlled study of zero-shot cross-embodiment transfer in VLA models for stationary tabletop manipulation with two-finger-gripper manipulators.
We distinguish strict from pretrain-exposed zero-shot transfer and evaluate appearance, gripper, arm, and full-embodiment shifts.
Our results show that local EEF-centered state-action representations, greater source-embodiment diversity under the studied comparisons, and auxiliary co-training improve strict transfer, while even limited target-embodiment exposure during pretraining substantially narrows the transfer gap.
These findings offer practical guidance within this controlled setting: explicitly report target-embodiment exposure, use local EEF-centered state-action representations, evaluate source-embodiment diversity as a controlled design variable, and incorporate auxiliary co-training.

%===============================================================================

\clearpage
% The acknowledgments are automatically included only in the final and preprint versions of the paper.
%\acknowledgments{If a paper is accepted, the final camera-ready version will (and probably should) include acknowledgments. All acknowledgments go at the end of the paper, including thanks to reviewers who gave useful comments, to colleagues who contributed to the ideas, and to funding agencies and corporate sponsors that provided financial support.}

%===============================================================================

\bibliography{references}  % .bib

@misc{lap,
      title={LAP: Language-Action Pre-Training Enables Zero-shot Cross-Embodiment Transfer}, 
      author={Lihan Zha and Asher J. Hancock and Mingtong Zhang and Tenny Yin and Yixuan Huang and Dhruv Shah and Allen Z. Ren and Anirudha Majumdar},
      year={2026},
      eprint={2602.10556},
      archivePrefix={arXiv},
      primaryClass={cs.RO},
      url={https://arxiv.org/abs/2602.10556}, 
}

@misc{lin2026systematiccotraining,
      title={A Systematic Study of Data Modalities and Strategies for Co-training Large Behavior Models for Robot Manipulation},
      author={Fanqi Lin and Kushal Arora and Jean Mercat and Haruki Nishimura and Paarth Shah and Chen Xu and Mengchao Zhang and Mark Zolotas and Maya Angeles and Owen Pfannenstiehl and Andrew Beaulieu and Jose Barreiros},
      year={2026},
      eprint={2602.01067},
      archivePrefix={arXiv},
      primaryClass={cs.RO},
      url={https://arxiv.org/abs/2602.01067},
}

@misc{xskill,
      title={XSkill: Cross Embodiment Skill Discovery}, 
      author={Mengda Xu and Zhenjia Xu and Cheng Chi and Manuela Veloso and Shuran Song},
      year={2023},
      eprint={2307.09955},
      archivePrefix={arXiv},
      primaryClass={cs.RO},
      url={https://arxiv.org/abs/2307.09955}, 
}

@misc{uniskill,
      title={UniSkill: Imitating Human Videos via Cross-Embodiment Skill Representations}, 
      author={Hanjung Kim and Jaehyun Kang and Hyolim Kang and Meedeum Cho and Seon Joo Kim and Youngwoon Lee},
      year={2025},
      eprint={2505.08787},
      archivePrefix={arXiv},
      primaryClass={cs.RO},
      url={https://arxiv.org/abs/2505.08787}, 
}

@misc{uniact,
      title={Universal Actions for Enhanced Embodied Foundation Models}, 
      author={Jinliang Zheng and Jianxiong Li and Dongxiu Liu and Yinan Zheng and Zhihao Wang and Zhonghong Ou and Yu Liu and Jingjing Liu and Ya-Qin Zhang and Xianyuan Zhan},
      year={2025},
      eprint={2501.10105},
      archivePrefix={arXiv},
      primaryClass={cs.RO},
      url={https://arxiv.org/abs/2501.10105}, 
}

@misc{ye2025latentactionpretraining,
      title={Latent Action Pretraining from Videos},
      author={Seonghyeon Ye and Joel Jang and Byeongguk Jeon and Sejune Joo and Jianwei Yang and Baolin Peng and Ajay Mandlekar and Reuben Tan and Yu-Wei Chao and Bill Yuchen Lin and Lars Liden and Kimin Lee and Jianfeng Gao and Luke Zettlemoyer and Dieter Fox and Minjoon Seo},
      year={2025},
      eprint={2410.11758},
      archivePrefix={arXiv},
      primaryClass={cs.RO},
      url={https://arxiv.org/abs/2410.11758},
}

@misc{xlvla,
      title={Cross-Hand Latent Representation for Vision-Language-Action Models}, 
      author={Guangqi Jiang and Yutong Liang and Jianglong Ye and Jia-Yang Huang and Changwei Jing and Rocky Duan and Pieter Abbeel and Xiaolong Wang and Xueyan Zou},
      year={2026},
      eprint={2603.10158},
      archivePrefix={arXiv},
      primaryClass={cs.RO},
      url={https://arxiv.org/abs/2603.10158}, 
}

@misc{motiontracks,
      title={Motion Tracks: A Unified Representation for Human-Robot Transfer in Few-Shot Imitation Learning}, 
      author={Juntao Ren and Priya Sundaresan and Dorsa Sadigh and Sanjiban Choudhury and Jeannette Bohg},
      year={2025},
      eprint={2501.06994},
      archivePrefix={arXiv},
      primaryClass={cs.RO},
      url={https://arxiv.org/abs/2501.06994}, 
}

@misc{cei,
      title={CEI: A Unified Interface for Cross-Embodiment Visuomotor Policy Learning in 3D Space}, 
      author={Tong Wu and Shoujie Li and Junhao Gong and Changqing Guo and Xingting Li and Shilong Mu and Wenbo Ding},
      year={2026},
      eprint={2601.09163},
      archivePrefix={arXiv},
      primaryClass={cs.RO},
      url={https://arxiv.org/abs/2601.09163}, 
}

@misc{ate,
      title={Align-Then-stEer: Adapting the Vision-Language Action Models through Unified Latent Guidance}, 
      author={Yang Zhang and Chenwei Wang and Ouyang Lu and Yuan Zhao and Yunfei Ge and Zhenglong Sun and Xiu Li and Chi Zhang and Chenjia Bai and Xuelong Li},
      year={2025},
      eprint={2509.02055},
      archivePrefix={arXiv},
      primaryClass={cs.RO},
      url={https://arxiv.org/abs/2509.02055}, 
}

@inproceedings{hpt,
      title={Scaling Proprioceptive-Visual Learning with Heterogeneous Pre-trained Transformers},
      author={Lirui Wang and Xinlei Chen and Jialiang Zhao and Kaiming He},
      booktitle={Advances in Neural Information Processing Systems},
      year={2024},
      eprint={2409.20537},
      archivePrefix={arXiv},
      primaryClass={cs.RO},
      url={https://arxiv.org/abs/2409.20537},
}

@misc{urma,
      title={One Policy to Run Them All: an End-to-end Learning Approach to Multi-Embodiment Locomotion}, 
      author={Nico Bohlinger and Grzegorz Czechmanowski and Maciej Krupka and Piotr Kicki and Krzysztof Walas and Jan Peters and Davide Tateo},
      year={2025},
      eprint={2409.06366},
      archivePrefix={arXiv},
      primaryClass={cs.RO},
      url={https://arxiv.org/abs/2409.06366}, 
}

@misc{xvla,
      title={X-VLA: Soft-Prompted Transformer as Scalable Cross-Embodiment Vision-Language-Action Model}, 
      author={Jinliang Zheng and Jianxiong Li and Zhihao Wang and Dongxiu Liu and Xirui Kang and Yuchun Feng and Yinan Zheng and Jiayin Zou and Yilun Chen and Jia Zeng and Ya-Qin Zhang and Jiangmiao Pang and Jingjing Liu and Tai Wang and Xianyuan Zhan},
      year={2025},
      eprint={2510.10274},
      archivePrefix={arXiv},
      primaryClass={cs.RO},
      url={https://arxiv.org/abs/2510.10274}, 
}

@misc{univla,
      title={UniVLA: Learning to Act Anywhere with Task-centric Latent Actions}, 
      author={Qingwen Bu and Yanting Yang and Jisong Cai and Shenyuan Gao and Guanghui Ren and Maoqing Yao and Ping Luo and Hongyang Li},
      year={2025},
      eprint={2505.06111},
      archivePrefix={arXiv},
      primaryClass={cs.RO},
      url={https://arxiv.org/abs/2505.06111}, 
}

@misc{anybody2025,
      title={AnyBody: A Benchmark Suite for Cross-Embodiment Manipulation},
      author={Meenal Parakh and Alexandre Kirchmeyer and Beining Han and Jia Deng},
      year={2025},
      eprint={2505.14986},
      archivePrefix={arXiv},
      primaryClass={cs.RO},
      url={https://arxiv.org/abs/2505.14986},
}

@misc{motif,
      title={MOTIF: Learning Action Motifs for Few-shot Cross-Embodiment Transfer}, 
      author={Heng Zhi and Wentao Tan and Lei Zhu and Fengling Li and Jingjing Li and Guoli Yang and Heng Tao Shen},
      year={2026},
      eprint={2602.13764},
      archivePrefix={arXiv},
      primaryClass={cs.RO},
      url={https://arxiv.org/abs/2602.13764}, 
}

@misc{egovla,
      title={EgoVLA: Learning Vision-Language-Action Models from Egocentric Human Videos}, 
      author={Ruihan Yang and Qinxi Yu and Yecheng Wu and Rui Yan and Borui Li and An-Chieh Cheng and Xueyan Zou and Yunhao Fang and Xuxin Cheng and Ri-Zhao Qiu and Hongxu Yin and Sifei Liu and Song Han and Yao Lu and Xiaolong Wang},
      year={2025},
      eprint={2507.12440},
      archivePrefix={arXiv},
      primaryClass={cs.RO},
      url={https://arxiv.org/abs/2507.12440}, 
}

@misc{dexvla,
      title={DexVLA: Vision-Language Model with Plug-In Diffusion Expert for General Robot Control}, 
      author={Junjie Wen and Yichen Zhu and Jinming Li and Zhibin Tang and Chaomin Shen and Feifei Feng},
      year={2025},
      eprint={2502.05855},
      archivePrefix={arXiv},
      primaryClass={cs.RO},
      url={https://arxiv.org/abs/2502.05855}, 
}

@article{legato,
      title={LEGATO: Cross-Embodiment Imitation Using a Grasping Tool},
      author={Mingyo Seo and H. Andy Park and Shenli Yuan and Yuke Zhu and Luis Sentis},
      journal={IEEE Robotics and Automation Letters},
      volume={10},
      number={3},
      pages={2854--2861},
      year={2025},
      doi={10.1109/LRA.2025.3535182},
      eprint={2411.03682},
      archivePrefix={arXiv},
      primaryClass={cs.RO},
      url={https://arxiv.org/abs/2411.03682},
}

@misc{mirage,
      title={Mirage: Cross-Embodiment Zero-Shot Policy Transfer with Cross-Painting},
      author={Lawrence Yunliang Chen and Kush Hari and Karthik Dharmarajan and Chenfeng Xu and Quan Vuong and Ken Goldberg},
      year={2024},
      eprint={2402.19249},
      archivePrefix={arXiv},
      primaryClass={cs.RO},
      url={https://arxiv.org/abs/2402.19249},
}

@misc{roviaug,
      title={RoVi-Aug: Robot and Viewpoint Augmentation for Cross-Embodiment Robot Learning},
      author={Lawrence Yunliang Chen and Chenfeng Xu and Karthik Dharmarajan and Muhammad Zubair Irshad and Richard Cheng and Kurt Keutzer and Masayoshi Tomizuka and Quan Vuong and Ken Goldberg},
      year={2024},
      eprint={2409.03403},
      archivePrefix={arXiv},
      primaryClass={cs.RO},
      url={https://arxiv.org/abs/2409.03403},
}

@misc{crossformer,
      title={Scaling Cross-Embodied Learning: One Policy for Manipulation, Navigation, Locomotion and Aviation},
      author={Ria Doshi and Homer Walke and Oier Mees and Sudeep Dasari and Sergey Levine},
      year={2024},
      eprint={2408.11812},
      archivePrefix={arXiv},
      primaryClass={cs.RO},
      url={https://arxiv.org/abs/2408.11812},
}

@inproceedings{robomind,
      title={RoboMIND: Benchmark on Multi-embodiment Intelligence Normative Data for Robot Manipulation},
      author={Kun Wu and Chengkai Hou and Jiaming Liu and Zhengping Che and Xiaozhu Ju and
              Zhuqin Yang and Meng Li and Yinuo Zhao and Zhiyuan Xu and Guang Yang and
              Shichao Fan and Xinhua Wang and Fei Liao and Zhen Zhao and Guangyu Li and Zhao Jin and
              Lecheng Wang and Jilei Mao and Ning Liu and Pei Ren and Qiang Zhang and Yaoxu Lyu and
              Mengzhen Liu and Jingyang He and Yulin Luo and Zeyu Gao and Chenxuan Li and
              Chenyang Gu and Yankai Fu and Di Wu and Xingyu Wang and Sixiang Chen and
              Zhenyu Wang and Pengju An and Siyuan Qian and Shanghang Zhang and Jian Tang},
      booktitle={Robotics: Science and Systems XXI},
      publisher={Robotics: Science and Systems Foundation},
      year={2025},
      month={June},
      doi={10.15607/RSS.2025.XXI.152},
      url={https://www.roboticsproceedings.org/rss21/p152.html},
      eprint={2412.13877},
      archivePrefix={arXiv},
      primaryClass={cs.RO},
}

@misc{diversityall2025,
      title={Is Diversity All You Need for Scalable Robotic Manipulation?}, 
      author={Modi Shi and Li Chen and Jin Chen and Yuxiang Lu and Chiming Liu and Guanghui Ren and Ping Luo and Di Huang and Maoqing Yao and Hongyang Li},
      year={2025},
      eprint={2507.06219},
      archivePrefix={arXiv},
      primaryClass={cs.RO},
      url={https://arxiv.org/abs/2507.06219}, 
}

@misc{embodimentscaling2025,
      title={Towards Embodiment Scaling Laws in Robot Locomotion}, 
      author={Bo Ai and Liu Dai and Nico Bohlinger and Dichen Li and Tongzhou Mu and Zhanxin Wu and K. Fay and Henrik I. Christensen and Jan Peters and Hao Su},
      year={2025},
      eprint={2505.05753},
      archivePrefix={arXiv},
      primaryClass={cs.RO},
      url={https://arxiv.org/abs/2505.05753}, 
}

@misc{shukor2025smolvla,
      title={SmolVLA: A Vision-Language-Action Model for Affordable and Efficient Robotics},
      author={Mustafa Shukor and Dana Aubakirova and Francesco Capuano and Pepijn Kooijmans and
              Steven Palma and Adil Zouitine and Michel Aractingi and Caroline Pascal and
              Martino Russi and Andres Marafioti and Simon Alibert and Matthieu Cord and
              Thomas Wolf and Remi Cadene},
      year={2025},
      eprint={2506.01844},
      archivePrefix={arXiv},
      primaryClass={cs.LG},
      url={https://arxiv.org/abs/2506.01844},
}

@inproceedings{wu2023gr1,
      title={Unleashing Large-Scale Video Generative Pre-training for Visual Robot Manipulation},
      author={Hongtao Wu and Ya Jing and Chilam Cheang and Guangzeng Chen and Jiafeng Xu and
              Xinghang Li and Minghuan Liu and Hang Li and Tao Kong},
      booktitle={International Conference on Learning Representations},
      year={2024},
      eprint={2312.13139},
      archivePrefix={arXiv},
      primaryClass={cs.RO},
      url={https://arxiv.org/abs/2312.13139},
}

@misc{li2023roboflamingo,
      title={Vision-Language Foundation Models as Effective Robot Imitators},
      author={Xinghang Li and Minghuan Liu and Hanbo Zhang and Cunjun Yu and Jie Xu and
              Hongtao Wu and Chilam Cheang and Ya Jing and Weinan Zhang and Huaping Liu and
              Hang Li and Tao Kong},
      year={2023},
      eprint={2311.01378},
      archivePrefix={arXiv},
      primaryClass={cs.RO},
      url={https://arxiv.org/abs/2311.01378},
}

@misc{nvidia2025gr00tn1,
      title={GR00T N1: An Open Foundation Model for Generalist Humanoid Robots},
      author={{NVIDIA} and Johan Bjorck and Fernando Castañeda and Nikita Cherniadev and
              Xingye Da and Runyu Ding and Linxi "Jim" Fan and Yu Fang and Dieter Fox and
              Fengyuan Hu and Spencer Huang and Joel Jang and Zhenyu Jiang and Jan Kautz and
              Kaushil Kundalia and Lawrence Lao and Zhiqi Li and Zongyu Lin and Kevin Lin and
              Guilin Liu and Edith Llontop and Loic Magne and Ajay Mandlekar and Avnish Narayan and
              Soroush Nasiriany and Scott Reed and You Liang Tan and Guanzhi Wang and Zu Wang and
              Jing Wang and Qi Wang and Jiannan Xiang and Yuqi Xie and Yinzhen Xu and Zhenjia Xu and
              Seonghyeon Ye and Zhiding Yu and Ao Zhang and Hao Zhang and Yizhou Zhao and
              Ruijie Zheng and Yuke Zhu},
      year={2025},
      eprint={2503.14734},
      archivePrefix={arXiv},
      primaryClass={cs.RO},
      url={https://arxiv.org/abs/2503.14734},
}

@InProceedings{zitkovich2023rt2,
  title = 	 {RT-2: Vision-Language-Action Models Transfer Web Knowledge to Robotic Control},
  author =       {Zitkovich, Brianna and Yu, Tianhe and Xu, Sichun and Xu, Peng and Xiao, Ted and Xia, Fei and Wu, Jialin and Wohlhart, Paul and Welker, Stefan and Wahid, Ayzaan and Vuong, Quan and Vanhoucke, Vincent and Tran, Huong and Soricut, Radu and Singh, Anikait and Singh, Jaspiar and Sermanet, Pierre and Sanketi, Pannag R. and Salazar, Grecia and Ryoo, Michael S. and Reymann, Krista and Rao, Kanishka and Pertsch, Karl and Mordatch, Igor and Michalewski, Henryk and Lu, Yao and Levine, Sergey and Lee, Lisa and Lee, Tsang-Wei Edward and Leal, Isabel and Kuang, Yuheng and Kalashnikov, Dmitry and Julian, Ryan and Joshi, Nikhil J. and Irpan, Alex and Ichter, Brian and Hsu, Jasmine and Herzog, Alexander and Hausman, Karol and Gopalakrishnan, Keerthana and Fu, Chuyuan and Florence, Pete and Finn, Chelsea and Dubey, Kumar Avinava and Driess, Danny and Ding, Tianli and Choromanski, Krzysztof Marcin and Chen, Xi and Chebotar, Yevgen and Carbajal, Justice and Brown, Noah and Brohan, Anthony and Arenas, Montserrat Gonzalez and Han, Kehang},
  booktitle = 	 {Proceedings of The 7th Conference on Robot Learning},
  pages = 	 {2165--2183},
  year = 	 {2023},
  editor = 	 {Tan, Jie and Toussaint, Marc and Darvish, Kourosh},
  volume = 	 {229},
  series = 	 {Proceedings of Machine Learning Research},
  month = 	 {06--09 Nov},
  publisher =    {PMLR},
  url = 	 {https://proceedings.mlr.press/v229/zitkovich23a.html}
}

@misc{liu2024robomambaefficientvisionlanguageactionmodel,
      title={RoboMamba: Efficient Vision-Language-Action Model for Robotic Reasoning and Manipulation}, 
      author={Jiaming Liu and Mengzhen Liu and Zhenyu Wang and Pengju An and Xiaoqi Li and Kaichen Zhou and Senqiao Yang and Renrui Zhang and Yandong Guo and Shanghang Zhang},
      year={2024},
      eprint={2406.04339},
      archivePrefix={arXiv},
      primaryClass={cs.CV},
      url={https://arxiv.org/abs/2406.04339}, 
}

@misc{physicalintelligence2025pi05,
      title={$\pi_{0.5}$: a Vision-Language-Action Model with Open-World Generalization},
      author={{Physical Intelligence} and Kevin Black and Noah Brown and James Darpinian and
              Karan Dhabalia and Danny Driess and Adnan Esmail and Michael Equi and Chelsea Finn and
              Niccolo Fusai and Manuel Y. Galliker and Dibya Ghosh and Lachy Groom and
              Karol Hausman and Brian Ichter and Szymon Jakubczak and Tim Jones and Liyiming Ke and
              Devin LeBlanc and Sergey Levine and Adrian Li-Bell and Mohith Mothukuri and
              Suraj Nair and Karl Pertsch and Allen Z. Ren and Lucy Xiaoyang Shi and Laura Smith and
              Jost Tobias Springenberg and Kyle Stachowicz and James Tanner and Quan Vuong and
              Homer Walke and Anna Walling and Haohuan Wang and Lili Yu and Ury Zhilinsky},
      year={2025},
      eprint={2504.16054},
      archivePrefix={arXiv},
      primaryClass={cs.LG},
      url={https://arxiv.org/abs/2504.16054},
}

@misc{lin2026onetwovlaunifiedvisionlanguageactionmodel,
      title={OneTwoVLA: A Unified Vision-Language-Action Model with Adaptive Reasoning}, 
      author={Fanqi Lin and Ruiqian Nai and Yingdong Hu and Jiacheng You and Junming Zhao and Yang Gao},
      year={2026},
      eprint={2505.11917},
      archivePrefix={arXiv},
      primaryClass={cs.RO},
      url={https://arxiv.org/abs/2505.11917}, 
}

@misc{qu2026eo1openunifiedembodied,
      title={EO-1: An Open Unified Embodied Foundation Model for General Robot Control}, 
      author={Delin Qu and Haoming Song and Qizhi Chen and Zhaoqing Chen and Xianqiang Gao and Dong Wang and Xinyi Ye and Qi Lv and Modi Shi and Guanghui Ren and Cheng Ruan and Maoqing Yao and Haoran Yang and Jiacheng Bao and Bin Zhao and Xuelong Li},
      year={2026},
      eprint={2508.21112},
      archivePrefix={arXiv},
      primaryClass={cs.RO},
      url={https://arxiv.org/abs/2508.21112}, 
}

@misc{chen2025internvlam1spatiallyguidedvisionlanguageaction,
      title={InternVLA-M1: A Spatially Guided Vision-Language-Action Framework for Generalist Robot Policy}, 
      author={Xinyi Chen and Yilun Chen and Yanwei Fu and Ning Gao and Jiaya Jia and Weiyang Jin and Hao Li and Yao Mu and Jiangmiao Pang and Yu Qiao and Yang Tian and Bin Wang and Bolun Wang and Fangjing Wang and Hanqing Wang and Tai Wang and Ziqin Wang and Xueyuan Wei and Chao Wu and Shuai Yang and Jinhui Ye and Junqiu Yu and Jia Zeng and Jingjing Zhang and Jinyu Zhang and Shi Zhang and Feng Zheng and Bowen Zhou and Yangkun Zhu},
      year={2025},
      eprint={2510.13778},
      archivePrefix={arXiv},
      primaryClass={cs.RO},
      url={https://arxiv.org/abs/2510.13778}, 
}

@misc{kim2024openvla,
      title={OpenVLA: An Open-Source Vision-Language-Action Model},
      author={Moo Jin Kim and Karl Pertsch and Siddharth Karamcheti and Ted Xiao and
              Ashwin Balakrishna and Suraj Nair and Rafael Rafailov and Ethan Foster and
              Grace Lam and Pannag Sanketi and Quan Vuong and Thomas Kollar and
              Benjamin Burchfiel and Russ Tedrake and Dorsa Sadigh and Sergey Levine and
              Percy Liang and Chelsea Finn},
      year={2024},
      eprint={2406.09246},
      archivePrefix={arXiv},
      primaryClass={cs.RO},
      url={https://arxiv.org/abs/2406.09246},
}

@misc{ecot,
      title={Robotic Control via Embodied Chain-of-Thought Reasoning}, 
      author={Michał Zawalski and William Chen and Karl Pertsch and Oier Mees and Chelsea Finn and Sergey Levine},
      year={2025},
      eprint={2407.08693},
      archivePrefix={arXiv},
      primaryClass={cs.RO},
      url={https://arxiv.org/abs/2407.08693}, 
}

@misc{deng2025graspvlagraspingfoundationmodel,
      title={GraspVLA: a Grasping Foundation Model Pre-trained on Billion-scale Synthetic Action Data}, 
      author={Shengliang Deng and Mi Yan and Songlin Wei and Haixin Ma and Yuxin Yang and Jiayi Chen and Zhiqi Zhang and Taoyu Yang and Xuheng Zhang and Wenhao Zhang and Heming Cui and Zhizheng Zhang and He Wang},
      year={2025},
      eprint={2505.03233},
      archivePrefix={arXiv},
      primaryClass={cs.RO},
      url={https://arxiv.org/abs/2505.03233}, 
}

@misc{maddukuri2025simandrealcotrainingsimplerecipe,
      title={Sim-and-Real Co-Training: A Simple Recipe for Vision-Based Robotic Manipulation}, 
      author={Abhiram Maddukuri and Zhenyu Jiang and Lawrence Yunliang Chen and Soroush Nasiriany and Yuqi Xie and Yu Fang and Wenqi Huang and Zu Wang and Zhenjia Xu and Nikita Chernyadev and Scott Reed and Ken Goldberg and Ajay Mandlekar and Linxi Fan and Yuke Zhu},
      year={2025},
      eprint={2503.24361},
      archivePrefix={arXiv},
      primaryClass={cs.RO},
      url={https://arxiv.org/abs/2503.24361}, 
}

@misc{yin2026geniesim30,
      title={Genie Sim 3.0 : A High-Fidelity Comprehensive Simulation Platform for Humanoid Robot}, 
      author={Chenghao Yin and Da Huang and Di Yang and Jichao Wang and Nanshu Zhao and Chen Xu and Wenjun Sun and Linjie Hou and Zhijun Li and Junhui Wu and Zhaobo Liu and Zhen Xiao and Sheng Zhang and Lei Bao and Rui Feng and Zhenquan Pang and Jiayu Li and Qian Wang and Maoqing Yao},
      year={2026},
      eprint={2601.02078},
      archivePrefix={arXiv},
      primaryClass={cs.RO},
      url={https://arxiv.org/abs/2601.02078}, 
}

@misc{interndataa1,
      title={InternData-A1: Pioneering High-Fidelity Synthetic Data for Pre-training Generalist Policy}, 
      author={Yang Tian and Yuyin Yang and Yiman Xie and Zetao Cai and Xu Shi and Ning Gao and Hangxu Liu and Xuekun Jiang and Zherui Qiu and Feng Yuan and Yaping Li and Ping Wang and Junhao Cai and Jia Zeng and Hao Dong and Jiangmiao Pang},
      year={2025},
      eprint={2511.16651},
      archivePrefix={arXiv},
      primaryClass={cs.RO},
      url={https://arxiv.org/abs/2511.16651}, 
}

@misc{locoformer,
      title={LocoFormer: Generalist Locomotion via Long-context Adaptation}, 
      author={Min Liu and Deepak Pathak and Ananye Agarwal},
      year={2025},
      eprint={2509.23745},
      archivePrefix={arXiv},
      primaryClass={cs.RO},
      url={https://arxiv.org/abs/2509.23745}, 
}

@misc{zhao2025needproprioceptivestatesvisuomotor,
      title={Do You Need Proprioceptive States in Visuomotor Policies?}, 
      author={Juntu Zhao and Wenbo Lu and Di Zhang and Yufeng Liu and Yushen Liang and Tianluo Zhang and Yifeng Cao and Junyuan Xie and Yingdong Hu and Shengjie Wang and Junliang Guo and Dequan Wang and Yang Gao},
      year={2025},
      eprint={2509.18644},
      archivePrefix={arXiv},
      primaryClass={cs.RO},
      url={https://arxiv.org/abs/2509.18644}, 
}

@misc{cotvla,
      title={CoT-VLA: Visual Chain-of-Thought Reasoning for Vision-Language-Action Models}, 
      author={Qingqing Zhao and Yao Lu and Moo Jin Kim and Zipeng Fu and Zhuoyang Zhang and Yecheng Wu and Zhaoshuo Li and Qianli Ma and Song Han and Chelsea Finn and Ankur Handa and Ming-Yu Liu and Donglai Xiang and Gordon Wetzstein and Tsung-Yi Lin},
      year={2025},
      eprint={2503.22020},
      archivePrefix={arXiv},
      primaryClass={cs.CV},
      url={https://arxiv.org/abs/2503.22020}, 
}

@misc{tracevla,
      title={TraceVLA: Visual Trace Prompting Enhances Spatial-Temporal Awareness for Generalist Robotic Policies}, 
      author={Ruijie Zheng and Yongyuan Liang and Shuaiyi Huang and Jianfeng Gao and Daumé III, Hal and Andrey Kolobov and Furong Huang and Jianwei Yang},
      year={2025},
      eprint={2412.10345},
      archivePrefix={arXiv},
      primaryClass={cs.RO},
      url={https://arxiv.org/abs/2412.10345}, 
}

@misc{kim2025openvlaoft,
      title={Fine-Tuning Vision-Language-Action Models: Optimizing Speed and Success},
      author={Moo Jin Kim and Chelsea Finn and Percy Liang},
      year={2025},
      eprint={2502.19645},
      archivePrefix={arXiv},
      primaryClass={cs.RO},
      url={https://arxiv.org/abs/2502.19645},
}

@misc{li2024cogact,
      title={CogACT: A Foundational Vision-Language-Action Model for Synergizing Cognition and Action in Robotic Manipulation},
      author={Qixiu Li and Yaobo Liang and Zeyu Wang and Lin Luo and Xi Chen and Mozheng Liao and
              Fangyun Wei and Yu Deng and Sicheng Xu and Yizhong Zhang and Xiaofan Wang and
              Bei Liu and Jianlong Fu and Jianmin Bao and Dong Chen and Yuanchun Shi and
              Jiaolong Yang and Baining Guo},
      year={2024},
      eprint={2411.19650},
      archivePrefix={arXiv},
      primaryClass={cs.RO},
      url={https://arxiv.org/abs/2411.19650},
}

@inproceedings{octo2024,
  title={Octo: An Open-Source Generalist Robot Policy},
  author={Ghosh, Dibya and Walke, Homer Rich and Pertsch, Karl and Black, Kevin and Mees, Oier and Dasari, Sudeep and Hejna, Joey and Kreiman, Tobias and Xu, Charles and Luo, Jianlan and Tan, You Liang and Chen, Lawrence Yunliang and Vuong, Quan and Xiao, Ted and Sanketi, Pannag R. and Sadigh, Dorsa and Finn, Chelsea and Levine, Sergey},
  booktitle={Proceedings of Robotics: Science and Systems},
  year={2024},
  address={Delft, Netherlands},
  month={July},
  doi={10.15607/RSS.2024.XX.090}
}

@inproceedings{openx2024rtx,
      title={Open X-Embodiment: Robotic Learning Datasets and RT-X Models},
      author={{Open X-Embodiment Collaboration} and Abby O'Neill and Abdul Rehman and Abhinav Gupta and
              Abhiram Maddukuri and Abhishek Gupta and Abhishek Padalkar and Abraham Lee and
              Acorn Pooley and Agrim Gupta and Ajay Mandlekar and Ajinkya Jain and Albert Tung and
              Alex Bewley and Alex Herzog and Alex Irpan and Alexander Khazatsky and Anant Rai and
              Anchit Gupta and Andrew Wang and Andrey Kolobov and Anikait Singh and Animesh Garg and
              Aniruddha Kembhavi and Annie Xie and Anthony Brohan and Antonin Raffin and
              Archit Sharma and Arefeh Yavary and Arhan Jain and Ashwin Balakrishna and Ayzaan Wahid and
              Ben Burgess-Limerick and Beomjoon Kim and Bernhard Schölkopf and Blake Wulfe and
              Brian Ichter and Cewu Lu and Charles Xu and Charlotte Le and Chelsea Finn and
              Chen Wang and Chenfeng Xu and Cheng Chi and Chenguang Huang and Christine Chan and
              Christopher Agia and Chuer Pan and Chuyuan Fu and Coline Devin and Danfei Xu and
              Daniel Morton and Danny Driess and Daphne Chen and Deepak Pathak and Dhruv Shah and
              Dieter Büchler and Dinesh Jayaraman and Dmitry Kalashnikov and Dorsa Sadigh and
              Edward Johns and Ethan Foster and Fangchen Liu and Federico Ceola and Fei Xia and
              Feiyu Zhao and Felipe Vieira Frujeri and Freek Stulp and Gaoyue Zhou and
              Gaurav S. Sukhatme and Gautam Salhotra and Ge Yan and Gilbert Feng and Giulio Schiavi and
              Glen Berseth and Gregory Kahn and Guangwen Yang and Guanzhi Wang and Hao Su and
              Hao-Shu Fang and Haochen Shi and Henghui Bao and Heni Ben Amor and
              Henrik I Christensen and Hiroki Furuta and Homanga Bharadhwaj and Homer Walke and
              Hongjie Fang and Huy Ha and Igor Mordatch and Ilija Radosavovic and Isabel Leal and
              Jacky Liang and Jad Abou-Chakra and Jaehyung Kim and Jaimyn Drake and Jan Peters and
              Jan Schneider and Jasmine Hsu and Jay Vakil and Jeannette Bohg and Jeffrey Bingham and
              Jeffrey Wu and Jensen Gao and Jiaheng Hu and Jiajun Wu and Jialin Wu and Jiankai Sun and
              Jianlan Luo and Jiayuan Gu and Jie Tan and Jihoon Oh and Jimmy Wu and Jingpei Lu and
              Jingyun Yang and Jitendra Malik and João Silvério and Joey Hejna and Jonathan Booher and
              Jonathan Tompson and Jonathan Yang and Jordi Salvador and Joseph J. Lim and
              Junhyek Han and Kaiyuan Wang and Kanishka Rao and Karl Pertsch and Karol Hausman and
              Keegan Go and Keerthana Gopalakrishnan and Ken Goldberg and Kendra Byrne and
              Kenneth Oslund and Kento Kawaharazuka and Kevin Black and Kevin Lin and Kevin Zhang and
              Kiana Ehsani and Kiran Lekkala and Kirsty Ellis and Krishan Rana and
              Krishnan Srinivasan and Kuan Fang and Kunal Pratap Singh and Kuo-Hao Zeng and
              Kyle Hatch and Kyle Hsu and Laurent Itti and Lawrence Yunliang Chen and Lerrel Pinto and
              Li Fei-Fei and Liam Tan and Linxi "Jim" Fan and Lionel Ott and Lisa Lee and Luca Weihs and
              Magnum Chen and Marion Lepert and Marius Memmel and Masayoshi Tomizuka and
              Masha Itkina and Mateo Guaman Castro and Max Spero and Maximilian Du and Michael Ahn and
              Michael C. Yip and Mingtong Zhang and Mingyu Ding and Minho Heo and
              Mohan Kumar Srirama and Mohit Sharma and Moo Jin Kim and Muhammad Zubair Irshad and
              Naoaki Kanazawa and Nicklas Hansen and Nicolas Heess and Nikhil J Joshi and
              Niko Suenderhauf and Ning Liu and Norman Di Palo and Nur Muhammad Mahi Shafiullah and
              Oier Mees and Oliver Kroemer and Osbert Bastani and Pannag R Sanketi and
              Patrick "Tree" Miller and Patrick Yin and Paul Wohlhart and Peng Xu and
              Peter David Fagan and Peter Mitrano and Pierre Sermanet and Pieter Abbeel and
              Priya Sundaresan and Qiuyu Chen and Quan Vuong and Rafael Rafailov and Ran Tian and
              Ria Doshi and Roberto Martín-Martín and Rohan Baijal and Rosario Scalise and
              Rose Hendrix and Roy Lin and Runjia Qian and Ruohan Zhang and Russell Mendonca and
              Rutav Shah and Ryan Hoque and Ryan Julian and Samuel Bustamante and Sean Kirmani and
              Sergey Levine and Shan Lin and Sherry Moore and Shikhar Bahl and Shivin Dass and
              Shubham Sonawani and Shubham Tulsiani and Shuran Song and Sichun Xu and
              Siddhant Haldar and Siddharth Karamcheti and Simeon Adebola and Simon Guist and
              Soroush Nasiriany and Stefan Schaal and Stefan Welker and Stephen Tian and
              Subramanian Ramamoorthy and Sudeep Dasari and Suneel Belkhale and Sungjae Park and
              Suraj Nair and Suvir Mirchandani and Takayuki Osa and Tanmay Gupta and Tatsuya Harada and
              Tatsuya Matsushima and Ted Xiao and Thomas Kollar and Tianhe Yu and Tianli Ding and
              Todor Davchev and Tony Z. Zhao and Travis Armstrong and Trevor Darrell and
              Trinity Chung and Vidhi Jain and Vikash Kumar and Vincent Vanhoucke and
              Vitor Guizilini and Wei Zhan and Wenxuan Zhou and Wolfram Burgard and Xi Chen and
              Xiangyu Chen and Xiaolong Wang and Xinghao Zhu and Xinyang Geng and Xiyuan Liu and
              Xu Liangwei and Xuanlin Li and Yansong Pang and Yao Lu and Yecheng Jason Ma and
              Yejin Kim and Yevgen Chebotar and Yifan Zhou and Yifeng Zhu and Yilin Wu and Ying Xu and
              Yixuan Wang and Yonatan Bisk and Yongqiang Dou and Yoonyoung Cho and Youngwoon Lee and
              Yuchen Cui and Yue Cao and Yueh-Hua Wu and Yujin Tang and Yuke Zhu and Yunchu Zhang and
              Yunfan Jiang and Yunshuang Li and Yunzhu Li and Yusuke Iwasawa and Yutaka Matsuo and
              Zehan Ma and Zhuo Xu and Zichen Jeff Cui and Zichen Zhang and Zipeng Fu and Zipeng Lin},
      booktitle={2024 IEEE International Conference on Robotics and Automation (ICRA)},
      pages={6892--6903},
      year={2024},
      doi={10.1109/ICRA57147.2024.10611477},
      eprint={2310.08864},
      archivePrefix={arXiv},
      primaryClass={cs.RO},
      url={https://arxiv.org/abs/2310.08864},
}

@inproceedings{yang2024pushing,
  title={Pushing the Limits of Cross-Embodiment Learning for Manipulation and Navigation},
  author={Yang, Jonathan H. and Glossop, Catherine and Bhorkar, Arjun and Shah, Dhruv and Vuong, Quan and Finn, Chelsea and Sadigh, Dorsa and Levine, Sergey},
  booktitle={Proceedings of Robotics: Science and Systems},
  year={2024},
  doi={10.15607/RSS.2024.XX.093}
}

@misc{khazatsky2025droid,
      title={DROID: A Large-Scale In-The-Wild Robot Manipulation Dataset}, 
      author={Alexander Khazatsky and Karl Pertsch and Suraj Nair and Ashwin Balakrishna and Sudeep Dasari and Siddharth Karamcheti and Soroush Nasiriany and Mohan Kumar Srirama and Lawrence Yunliang Chen and Kirsty Ellis and Peter David Fagan and Joey Hejna and Masha Itkina and Marion Lepert and Yecheng Jason Ma and Patrick Tree Miller and Jimmy Wu and Suneel Belkhale and Shivin Dass and Huy Ha and Arhan Jain and Abraham Lee and Youngwoon Lee and Marius Memmel and Sungjae Park and Ilija Radosavovic and Kaiyuan Wang and Albert Zhan and Kevin Black and Cheng Chi and Kyle Beltran Hatch and Shan Lin and Jingpei Lu and Jean Mercat and Abdul Rehman and Pannag R Sanketi and Archit Sharma and Cody Simpson and Quan Vuong and Homer Rich Walke and Blake Wulfe and Ted Xiao and Jonathan Heewon Yang and Arefeh Yavary and Tony Z. Zhao and Christopher Agia and Rohan Baijal and Mateo Guaman Castro and Daphne Chen and Qiuyu Chen and Trinity Chung and Jaimyn Drake and Ethan Paul Foster and Jensen Gao and Vitor Guizilini and David Antonio Herrera and Minho Heo and Kyle Hsu and Jiaheng Hu and Muhammad Zubair Irshad and Donovon Jackson and Charlotte Le and Yunshuang Li and Kevin Lin and Roy Lin and Zehan Ma and Abhiram Maddukuri and Suvir Mirchandani and Daniel Morton and Tony Nguyen and Abigail O'Neill and Rosario Scalise and Derick Seale and Victor Son and Stephen Tian and Emi Tran and Andrew E. Wang and Yilin Wu and Annie Xie and Jingyun Yang and Patrick Yin and Yunchu Zhang and Osbert Bastani and Glen Berseth and Jeannette Bohg and Ken Goldberg and Abhinav Gupta and Abhishek Gupta and Dinesh Jayaraman and Joseph J Lim and Jitendra Malik and Roberto Martín-Martín and Subramanian Ramamoorthy and Dorsa Sadigh and Shuran Song and Jiajun Wu and Michael C. Yip and Yuke Zhu and Thomas Kollar and Sergey Levine and Chelsea Finn},
      year={2024},
      eprint={2403.12945},
      archivePrefix={arXiv},
      primaryClass={cs.RO},
      url={https://arxiv.org/abs/2403.12945}, 
}

@misc{fang2023rh20t,
      title={RH20T: A Comprehensive Robotic Dataset for Learning Diverse Skills in One-Shot}, 
      author={Hao-Shu Fang and Hongjie Fang and Zhenyu Tang and Jirong Liu and Chenxi Wang and Junbo Wang and Haoyi Zhu and Cewu Lu},
      year={2023},
      eprint={2307.00595},
      archivePrefix={arXiv},
      primaryClass={cs.RO},
      url={https://arxiv.org/abs/2307.00595}, 
}

@misc{brohan2022rt1,
      title={RT-1: Robotics Transformer for Real-World Control at Scale},
      author={Anthony Brohan and Noah Brown and Justice Carbajal and Yevgen Chebotar and Joseph Dabis and Chelsea Finn and Keerthana Gopalakrishnan and Karol Hausman and Alex Herzog and Jasmine Hsu and Julian Ibarz and Brian Ichter and Alex Irpan and Tomas Jackson and Sally Jesmonth and Nikhil J Joshi and Ryan Julian and Dmitry Kalashnikov and Yuheng Kuang and Isabel Leal and Kuang-Huei Lee and Sergey Levine and Yao Lu and Utsav Malla and Deeksha Manjunath and Igor Mordatch and Ofir Nachum and Carolina Parada and Jodilyn Peralta and Emily Perez and Karl Pertsch and Jornell Quiambao and Kanishka Rao and Michael Ryoo and Grecia Salazar and Pannag Sanketi and Kevin Sayed and Jaspiar Singh and Sumedh Sontakke and Austin Stone and Clayton Tan and Huong Tran and Vincent Vanhoucke and Steve Vega and Quan Vuong and Fei Xia and Ted Xiao and Peng Xu and Sichun Xu and Tianhe Yu and Brianna Zitkovich},
      year={2022},
      eprint={2212.06817},
      archivePrefix={arXiv},
      primaryClass={cs.RO},
      url={https://arxiv.org/abs/2212.06817},
}

@inproceedings{black2024pi0, 
    AUTHOR    = {Kevin Black AND Noah Brown AND Danny Driess AND Adnan Esmail AND Michael Robert Equi AND Chelsea Finn AND Niccolo Fusai AND Lachy Groom AND Karol Hausman AND Brian Ichter AND Szymon Jakubczak AND Tim Jones AND Liyiming Ke AND Sergey Levine AND Adrian Li-Bell AND Mohith Mothukuri AND Suraj Nair AND Karl Pertsch AND Lucy Xiaoyang Shi AND Laura Smith AND James Tanner AND Quan Vuong AND Anna Walling AND Haohuan Wang AND Ury Zhilinsky}, 
    TITLE     = {{$\pi_0$: A Vision-Language-Action Flow Model for General Robot Control}}, 
    BOOKTITLE = {Proceedings of Robotics: Science and Systems}, 
    YEAR      = {2025}, 
    ADDRESS   = {Los Angeles, CA, USA}, 
    MONTH     = {June}, 
    DOI       = {10.15607/RSS.2025.XXI.010} 
}

@misc{walke2024bridgedatav2,
      title={BridgeData V2: A Dataset for Robot Learning at Scale}, 
      author={Homer Walke and Kevin Black and Abraham Lee and Moo Jin Kim and Max Du and Chongyi Zheng and Tony Zhao and Philippe Hansen-Estruch and Quan Vuong and Andre He and Vivek Myers and Kuan Fang and Chelsea Finn and Sergey Levine},
      year={2023},
      eprint={2308.12952},
      archivePrefix={arXiv},
      primaryClass={cs.RO},
      url={https://arxiv.org/abs/2308.12952}, 
}

@misc{agibotworld2025,
      title={AgiBot World Colosseo: A Large-scale Manipulation Platform for Scalable and Intelligent Embodied Systems}, 
      author={{AgiBot-World-Contributors} and Qingwen Bu and Jisong Cai and Li Chen and Xiuqi Cui and Yan Ding and Siyuan Feng and Shenyuan Gao and Xindong He and Xuan Hu and Xu Huang and Shu Jiang and Yuxin Jiang and Cheng Jing and Hongyang Li and Jialu Li and Chiming Liu and Yi Liu and Yuxiang Lu and Jianlan Luo and Ping Luo and Yao Mu and Yuehan Niu and Yixuan Pan and Jiangmiao Pang and Yu Qiao and Guanghui Ren and Cheng Ruan and Jiaqi Shan and Yongjian Shen and Chengshi Shi and Mingkang Shi and Modi Shi and Chonghao Sima and Jianheng Song and Huijie Wang and Wenhao Wang and Dafeng Wei and Chengen Xie and Guo Xu and Junchi Yan and Cunbiao Yang and Lei Yang and Shukai Yang and Maoqing Yao and Jia Zeng and Chi Zhang and Qinglin Zhang and Bin Zhao and Chengyue Zhao and Jiaqi Zhao and Jianchao Zhu},
      year={2025},
      eprint={2503.06669},
      archivePrefix={arXiv},
      primaryClass={cs.RO},
      url={https://arxiv.org/abs/2503.06669}, 
}

@misc{ario2024,
      title={All Robots in One: A New Standard and Unified Dataset for Versatile, General-Purpose Embodied Agents}, 
      author={Zhiqiang Wang and Hao Zheng and Yunshuang Nie and Wenjun Xu and Qingwei Wang and Hua Ye and Zhe Li and Kaidong Zhang and Xuewen Cheng and Wanxi Dong and Chang Cai and Liang Lin and Feng Zheng and Xiaodan Liang},
      year={2024},
      eprint={2408.10899},
      archivePrefix={arXiv},
      primaryClass={cs.RO},
      url={https://arxiv.org/abs/2408.10899}, 
}

@misc{spatialvla2025,
      title={SpatialVLA: Exploring Spatial Representations for Visual-Language-Action Model}, 
      author={Delin Qu and Haoming Song and Qizhi Chen and Yuanqi Yao and Xinyi Ye and Yan Ding and Zhigang Wang and JiaYuan Gu and Bin Zhao and Dong Wang and Xuelong Li},
      year={2025},
      eprint={2501.15830},
      archivePrefix={arXiv},
      primaryClass={cs.RO},
      url={https://arxiv.org/abs/2501.15830}, 
}

@misc{liu2024rdt1b,
      title={RDT-1B: a Diffusion Foundation Model for Bimanual Manipulation}, 
      author={Songming Liu and Lingxuan Wu and Bangguo Li and Hengkai Tan and Huayu Chen and Zhengyi Wang and Ke Xu and Hang Su and Jun Zhu},
      year={2025},
      eprint={2410.07864},
      archivePrefix={arXiv},
      primaryClass={cs.RO},
      url={https://arxiv.org/abs/2410.07864}, 
}

@misc{shadow2025,
      title={SHADOW: Leveraging Segmentation Masks for Cross-Embodiment Policy Transfer},
      author={Marion Lepert and Ria Doshi and Jeannette Bohg},
      year={2025},
      eprint={2503.00774},
      archivePrefix={arXiv},
      primaryClass={cs.RO},
      url={https://arxiv.org/abs/2503.00774},
}

@misc{lepert2025phantom,
      title={Phantom: Training Robots Without Robots Using Only Human Videos},
      author={Marion Lepert and Jiaying Fang and Jeannette Bohg},
      year={2025},
      eprint={2503.00779},
      archivePrefix={arXiv},
      primaryClass={cs.RO},
      url={https://arxiv.org/abs/2503.00779},
}

@misc{wang2024crossembodimentlatent,
      title={Cross-Embodiment Robot Manipulation Skill Transfer using Latent Space Alignment},
      author={Tianyu Wang and Dwait Bhatt and Xiaolong Wang and Nikolay Atanasov},
      year={2024},
      eprint={2406.01968},
      archivePrefix={arXiv},
      primaryClass={cs.RO},
      url={https://arxiv.org/abs/2406.01968},
}

@misc{bauer2025latentactiondiffusion,
      title={Latent Action Diffusion for Cross-Embodiment Manipulation},
      author={Erik Bauer and Elvis Nava and Robert K. Katzschmann},
      year={2025},
      eprint={2506.14608},
      archivePrefix={arXiv},
      primaryClass={cs.RO},
      url={https://arxiv.org/abs/2506.14608},
}

@misc{mu2026opfa,
      title={One-Policy-Fits-All: Geometry-Aware Action Latents for Cross-Embodiment Manipulation},
      author={Juncheng Mu and Sizhe Yang and Hojin Bae and Feiyu Jia and Qingwei Ben and Boyi Li and Huazhe Xu and Jiangmiao Pang},
      year={2026},
      eprint={2603.14522},
      archivePrefix={arXiv},
      primaryClass={cs.RO},
      url={https://arxiv.org/abs/2603.14522},
}

@inproceedings{yuan2024crossdex,
      title={Cross-Embodiment Dexterous Grasping with Reinforcement Learning},
      author={Haoqi Yuan and Bohan Zhou and Yuhui Fu and Zongqing Lu},
      booktitle={International Conference on Learning Representations},
      year={2025},
      url={https://arxiv.org/abs/2410.02479},
}

@misc{yan2026unifiedlatent,
      title={Learning a Unified Latent Space for Cross-Embodiment Robot Control},
      author={Yashuai Yan and Dongheui Lee},
      year={2026},
      eprint={2601.15419},
      archivePrefix={arXiv},
      primaryClass={cs.RO},
      url={https://arxiv.org/abs/2601.15419},
}

@misc{capo2026,
      title={Learning Adaptive Cross-Embodiment Visuomotor Policy with Contrastive Prompt Orchestration}, 
      author={Yuhang Zhang and Chao Yan and Jiaxi Yu and Jiaping Xiao and Mir Feroskhan},
      year={2026},
      eprint={2602.01040},
      archivePrefix={arXiv},
      primaryClass={cs.RO},
      url={https://arxiv.org/abs/2602.01040}, 
}

@misc{intelligence2026pi07,
      title={${\pi}_{0.7}$: a Steerable Generalist Robotic Foundation Model with Emergent Capabilities}, 
      author={{Physical Intelligence} and Bo Ai and Ali Amin and Raichelle Aniceto and Ashwin Balakrishna and Greg Balke and Kevin Black and George Bokinsky and Shihao Cao and Thomas Charbonnier and Vedant Choudhary and Foster Collins and Ken Conley and Grace Connors and James Darpinian and Karan Dhabalia and Maitrayee Dhaka and Jared DiCarlo and Danny Driess and Michael Equi and Adnan Esmail and Yunhao Fang and Chelsea Finn and Catherine Glossop and Thomas Godden and Ivan Goryachev and Lachlan Groom and Haroun Habeeb and Hunter Hancock and Karol Hausman and Gashon Hussein and Victor Hwang and Brian Ichter and Connor Jacobsen and Szymon Jakubczak and Rowan Jen and Tim Jones and Gregg Kammerer and Ben Katz and Liyiming Ke and Mairbek Khadikov and Chandra Kuchi and Marinda Lamb and Devin LeBlanc and Brendon LeCount and Sergey Levine and Xinyu Li and Adrian Li-Bell and Vladislav Lialin and Zhonglin Liang and Wallace Lim and Yao Lu and Enyu Luo and Vishnu Mano and Nandan Marwaha and Aikys Mongush and Liam Murphy and Suraj Nair and Tyler Patterson and Karl Pertsch and Allen Z. Ren and Gavin Schelske and Charvi Sharma and Baifeng Shi and Lucy Xiaoyang Shi and Laura Smith and Jost Tobias Springenberg and Kyle Stachowicz and Will Stoeckle and Jiaming Tang and Jimmy Tanner and Shalom Tekeste and Marcel Torne and Kyle Vedder and Quan Vuong and Anna Walling and Haohuan Wang and Jason Wang and XuDong Wang and Chris Whalen and Samuel Whitmore and Blake Williams and Charles Xu and Sukwon Yoo and Lili Yu and Wuming Zhang and Zhuoyang Zhang and Ury Zhilinsky},
      year={2026},
      eprint={2604.15483},
      archivePrefix={arXiv},
      primaryClass={cs.LG},
      url={https://arxiv.org/abs/2604.15483}, 
}

@misc{luo2026being,
      title={{Being-H0.5}: Scaling Human-Centric Robot Learning for Cross-Embodiment Generalization}, 
      author={Hao Luo and Ye Wang and Wanpeng Zhang and Sipeng Zheng and Ziheng Xi and Chaoyi Xu and Haiweng Xu and Haoqi Yuan and Chi Zhang and Yiqing Wang and Yicheng Feng and Zongqing Lu},
      year={2026},
      eprint={2601.12993},
      archivePrefix={arXiv},
      primaryClass={cs.RO},
      url={https://arxiv.org/abs/2601.12993}, 
}

@misc{yuan2026qwenrobotmanip,
      title={Qwen-RobotManip Technical Report: Alignment Unlocks Scale for Robotic Manipulation Foundation Models}, 
      author={Haoqi Yuan and Zhixuan Liang and Anzhe Chen and Ye Wang and Haoyang Li and Pei Lin and Yiyang Huang and Zixing Lei and Tong Zhang and Jiazhao Zhang and Jie Zhang and Jingyang Fan and Gengze Zhou and Qihang Peng and Chenxu Lv and Xiaoyue Chen and An Yang and Fei Huang and Junyang Lin and Dayiheng Liu and Jingren Zhou and Chenfei Wu and Xiong-Hui Chen},
      year={2026},
      eprint={2606.17846},
      archivePrefix={arXiv},
      primaryClass={cs.RO},
      url={https://arxiv.org/abs/2606.17846}, 
}

@misc{piseno2026cloak,
      title={Cloak: Zero-Shot Cross-Embodiment Manipulation by Masking the End-Effector from the VLA}, 
      author={Michael Piseno and Guy Tevet and C. Karen Liu},
      year={2026},
      eprint={2606.22836},
      archivePrefix={arXiv},
      primaryClass={cs.RO},
      url={https://arxiv.org/abs/2606.22836}, 
}

@misc{liu2026rdt2,
      title={RDT2: Exploring the Scaling Limit of UMI Data Towards Zero-Shot Cross-Embodiment Generalization}, 
      author={Songming Liu and Bangguo Li and Kai Ma and Lingxuan Wu and Hengkai Tan and Xiao Ouyang and Hang Su and Jun Zhu},
      year={2026},
      eprint={2602.03310},
      archivePrefix={arXiv},
      primaryClass={cs.RO},
      url={https://arxiv.org/abs/2602.03310}, 
}

@misc{geminiroboticsteam2025geminirobotics15,
      title={Gemini Robotics 1.5: Pushing the Frontier of Generalist Robots with Advanced Embodied Reasoning, Thinking, and Motion Transfer}, 
      author={{Gemini Robotics Team} and Abbas Abdolmaleki and Saminda Abeyruwan and Joshua Ainslie and Jean-Baptiste Alayrac and Montserrat Gonzalez Arenas and Ashwin Balakrishna and Nathan Batchelor and Alex Bewley and Jeff Bingham and Michael Bloesch and Konstantinos Bousmalis and Philemon Brakel and Anthony Brohan and Thomas Buschmann and Arunkumar Byravan and Serkan Cabi and Ken Caluwaerts and Federico Casarini and Christine Chan and Oscar Chang and London Chappellet-Volpini and Jose Enrique Chen and Xi Chen and Hao-Tien Lewis Chiang and Krzysztof Choromanski and Adrian Collister and David B. D'Ambrosio and Sudeep Dasari and Todor Davchev and Meet Kirankumar Dave and Coline Devin and Norman Di Palo and Tianli Ding and Carl Doersch and Adil Dostmohamed and Yilun Du and Debidatta Dwibedi and Sathish Thoppay Egambaram and Michael Elabd and Tom Erez and Xiaolin Fang and Claudio Fantacci and Cody Fong and Erik Frey and Chuyuan Fu and Ruiqi Gao and Marissa Giustina and Keerthana Gopalakrishnan and Laura Graesser and Oliver Groth and Agrim Gupta and Roland Hafner and Steven Hansen and Leonard Hasenclever and Sam Haves and Nicolas Heess and Brandon Hernaez and Alex Hofer and Jasmine Hsu and Lu Huang and Sandy H. Huang and Atil Iscen and Mithun George Jacob and Deepali Jain and Sally Jesmonth and Abhishek Jindal and Ryan Julian and Dmitry Kalashnikov and M. Emre Karagozler and Stefani Karp and Matija Kecman and J. Chase Kew and Donnie Kim and Frank Kim and Junkyung Kim and Thomas Kipf and Sean Kirmani and Ksenia Konyushkova and Li Yang Ku and Yuheng Kuang and Thomas Lampe and Antoine Laurens and Tuan Anh Le and Isabel Leal and Alex X. Lee and Tsang-Wei Edward Lee and Guy Lever and Jacky Liang and Li-Heng Lin and Fangchen Liu and Shangbang Long and Caden Lu and Sharath Maddineni and Anirudha Majumdar and Kevis-Kokitsi Maninis and Andrew Marmon and Sergio Martinez and Assaf Hurwitz Michaely and Niko Milonopoulos and Joss Moore and Robert Moreno and Michael Neunert and Francesco Nori and Joy Ortiz and Kenneth Oslund and Carolina Parada and Emilio Parisotto and Amaris Paryag and Acorn Pooley and Thomas Power and Alessio Quaglino and Haroon Qureshi and Rajkumar Vasudeva Raju and Helen Ran and Dushyant Rao and Kanishka Rao and Isaac Reid and David Rendleman and Krista Reymann and Miguel Rivas and Francesco Romano and Yulia Rubanova and Peter Pastor Sampedro and Pannag R Sanketi and Dhruv Shah and Mohit Sharma and Kathryn Shea and Mohit Shridhar and Charles Shu and Vikas Sindhwani and Sumeet Singh and Radu Soricut and Rachel Sterneck and Ian Storz and Razvan Surdulescu and Jie Tan and Jonathan Tompson and Saran Tunyasuvunakool and Jake Varley and Grace Vesom and Giulia Vezzani and Maria Bauza Villalonga and Oriol Vinyals and René Wagner and Ayzaan Wahid and Stefan Welker and Paul Wohlhart and Chengda Wu and Markus Wulfmeier and Fei Xia and Ted Xiao and Annie Xie and Jinyu Xie and Peng Xu and Sichun Xu and Ying Xu and Zhuo Xu and Jimmy Yan and Sherry Yang and Skye Yang and Yuxiang Yang and Hiu Hong Yu and Wenhao Yu and Wentao Yuan and Yuan Yuan and Jingwei Zhang and Tingnan Zhang and Zhiyuan Zhang and Allan Zhou and Guangyao Zhou and Yuxiang Zhou},
      year={2025},
      eprint={2510.03342},
      archivePrefix={arXiv},
      primaryClass={cs.RO},
      url={https://arxiv.org/abs/2510.03342}, 
}

\clearpage
\appendix
\tightsection{Formal Definitions of the Two Zero-Shot Protocols}

\begin{table*}[h]
\tightfloatbefore
\centering
\small
\setlength{\tabcolsep}{3pt}
\renewcommand{\arraystretch}{1.25}
\caption{Formal distinction between the two zero-shot cross-embodiment transfer protocols.}
\label{tab:supp_zero_shot_protocols}
\begin{tabular}{@{}p{0.23\textwidth}p{0.25\textwidth}p{0.27\textwidth}p{0.11\textwidth}@{}}
\hline
\rule{0pt}{2.6ex}\textbf{Protocol}
& \rule{0pt}{2.6ex}\textbf{Pre-train data}
& \rule{0pt}{2.6ex}\textbf{Post-train data}
& \rule{0pt}{2.6ex}\textbf{Eval.} \\
\hline
\begin{tabular}[t]{@{}l@{}}
\textbf{Strict zero-shot}
\end{tabular}
&
\begin{tabular}[t]{@{}l@{}}
$\mathcal{D}_{\mathrm{pre}}=\mathcal{D}(\mathcal{E}_{\mathrm{pre}},\mathcal{T}_{\mathrm{pre}})$\\
\boldmath$e_t\notin\mathcal{E}_{\mathrm{pre}}$\\
$\mathcal{T}_{\mathrm{pre}}\cap\mathcal{T}_{\mathrm{post}}=\varnothing$
\end{tabular}
&
\begin{tabular}[t]{@{}l@{}}
$\mathcal{D}_{\mathrm{post}}=\mathcal{D}(\{e_s\},\mathcal{T}_{\mathrm{post}})$\\
$e_s\neq e_t$
\end{tabular}
&
\begin{tabular}[t]{@{}l@{}}
emb.: $e_t$\\
task: $\mathcal{T}_{\mathrm{post}}$
\end{tabular}
\\[0.3em]
\hline
\rule{0pt}{2.6ex}
\begin{tabular}[t]{@{}l@{}}
\textbf{Pretrain-}\\
\textbf{exposed zero-shot}
\end{tabular}
&
\begin{tabular}[t]{@{}l@{}}
$\mathcal{D}_{\mathrm{pre}}=\mathcal{D}(\mathcal{E}_{\mathrm{pre}},\mathcal{T}_{\mathrm{pre}})$\\
\boldmath$e_t\in\mathcal{E}_{\mathrm{pre}}$\\
$\mathcal{T}_{\mathrm{pre}}\cap\mathcal{T}_{\mathrm{post}}=\varnothing$
\end{tabular}
&
\begin{tabular}[t]{@{}l@{}}
$\mathcal{D}_{\mathrm{post}}=\mathcal{D}(\{e_s\},\mathcal{T}_{\mathrm{post}})$\\
$e_s\neq e_t$
\end{tabular}
&
\begin{tabular}[t]{@{}l@{}}
emb.: $e_t$\\
task: $\mathcal{T}_{\mathrm{post}}$
\end{tabular}
\\
\hline
\end{tabular}
\tightfloatafter
\end{table*}

\autoref{tab:supp_zero_shot_protocols} separates the two protocols by whether the target embodiment appears in the pretraining embodiment set.
Both protocols use the same post-training data, $\mathcal{D}_{\mathrm{post}}=\mathcal{D}(\{e_s\},\mathcal{T}_{\mathrm{post}})$ with $e_s\neq e_t$, and both evaluate on the target embodiment $e_t$ for downstream tasks $\mathcal{T}_{\mathrm{post}}$.
In strict zero-shot transfer, $e_t\notin\mathcal{E}_{\mathrm{pre}}$, so evaluation measures transfer to a robot absent from all training data.
In pretrain-exposed zero-shot transfer, $e_t\in\mathcal{E}_{\mathrm{pre}}$, but pretraining tasks remain disjoint from downstream tasks and post-training still uses only the source embodiment.
Thus the evaluation remains zero-shot with respect to target-embodiment downstream demonstrations, because $\mathcal{D}(\{e_t\},\mathcal{T}_{\mathrm{post}})$ is never used for training, but it is not strict zero-shot because the policy has already observed the target embodiment during pretraining.

\tightsection{Per-RQ Experimental Design Matrix}

\autoref{tab:supp_per_rq_design_matrix} summarizes the experimental configuration used for each research question.
The purpose of this matrix is to make explicit which factor is isolated in each comparison, which zero-shot protocol is used, what pretraining data enters the model, and whether the evidence comes from simulation, real-world evaluation, or both.
Across all RQs, post-training uses the same source-embodiment downstream data, and the policy backbone, visual inputs, action horizon, optimization schedule, downstream task set, source post-training embodiment, and evaluation scenes are held fixed unless explicitly varied.

\begin{table*}[h]
\tightfloatbefore
\centering
\setlength{\tabcolsep}{2pt}
\renewcommand{\arraystretch}{1.14}
\caption{Per-RQ experimental design matrix. The table summarizes the protocol, pretraining data, representation choice, and evaluation platform used to isolate each experimental factor.}
\label{tab:supp_per_rq_design_matrix}
\begin{tabular}{@{}>{\raggedright\arraybackslash}p{0.055\textwidth}
                >{\raggedright\arraybackslash}p{0.15\textwidth}
                >{\raggedright\arraybackslash}p{0.15\textwidth}
                >{\raggedright\arraybackslash}p{0.27\textwidth}
                >{\raggedright\arraybackslash}p{0.20\textwidth}
                >{\raggedright\arraybackslash}p{0.105\textwidth}@{}}
\hline
\rule{0pt}{2.6ex}\textbf{RQ}
& \rule{0pt}{2.6ex}\textbf{Factor studied}
& \rule{0pt}{2.6ex}\textbf{Zero-shot protocol}
& \rule{0pt}{2.6ex}\textbf{Pretraining data}
& \rule{0pt}{2.6ex}\textbf{Representation}
& \rule{0pt}{2.6ex}\textbf{Eval.} \\
\hline
RQ1
& State-action representation
& Strict zero-shot
& 512 non-target source embodiments
& Varied across four designs
& Sim. + real \\
\hline
RQ2
& Source-embodiment diversity
& Strict zero-shot
& Primary: 1/8/512 sources, fixed total budget. Complementary: 32/128/512 sources, fixed per-embodiment budget.
& Fixed to best-average RQ1
& Sim. + real \\
\hline
RQ3
& Auxiliary co-training objective
& Strict zero-shot
& 512 non-target source embodiments with auxiliary objectives
& Fixed to best-average RQ1
& Sim. \\
\hline
RQ4
& Target-embodiment exposure
& Pretrain-exposed zero-shot
& Fixed-size pretraining pool with controlled target-embodiment replacement
& Fixed to best-average RQ1
& Sim. \\
\hline
\end{tabular}%
\tightfloatafter
\end{table*}

\tightsection{Detailed Experimental Results}
\label{sec:supp_detailed_experimental_results}

\begin{table*}[htbp]
\tightfloatbefore
\centering
\scriptsize
\setlength{\tabcolsep}{2.8pt}
\renewcommand{\arraystretch}{1.08}
\caption{Detailed RQ1 simulation task progress for serving sausages. Rows report individual robots, while columns vary the state/action representation. Each entry reports run-1/ run-2/ run-3 task progress (\%).}
\label{tab:supp_rq1_sausage}
\resizebox{\textwidth}{!}{%
\begin{tabular}{lcccc}
\hline
\rule{0pt}{2.6ex}\textbf{Target embodiment} & \textbf{Abs. EEF / World-Delta} & \textbf{Abs. EEF / EEF-Delta} & \textbf{EEF-Delta / World-Delta} & \textbf{EEF-Delta / EEF-Delta} \\
\hline
Franka & 80.8/ 83.8/ 79.8 & 89.9/ 82.8/ 89.9 & 85.9/ 80.8/ 86.9 & 84.8/ 91.9/ 90.9 \\
Franka+logo & 83.8/ 79.8/ 82.8 & 83.8/ 90.9/ 93.9 & 86.9/ 85.9/ 86.9 & 88.9/ 85.9/ 86.9 \\
Franka+green & 84.8/ 83.8/ 82.8 & 87.9/ 88.9/ 89.9 & 81.8/ 82.8/ 86.9 & 85.9/ 91.9/ 86.9 \\
Franka+UMI & 79.8/ 78.8/ 77.8 & 83.8/ 90.9/ 83.8 & 70.7/ 80.8/ 76.8 & 74.8/ 81.8/ 86.9 \\
UR5e+Franka & 65.7/ 68.7/ 70.7 & 70.7/ 69.7/ 70.7 & 76.8/ 81.8/ 74.8 & 78.8/ 81.8/ 75.8 \\
Google+Franka & 0.0/ 0.0/ 0.0 & 2.0/ 1.0/ 1.0 & 54.5/ 53.5/ 53.5 & 50.5/ 58.6/ 56.6 \\
UR5e+UMI & 69.7/ 60.6/ 58.6 & 69.7/ 76.8/ 75.8 & 69.7/ 62.7/ 59.6 & 69.7/ 69.7/ 69.7 \\
GoogleRobot & 1.0/ 1.0/ 0.0 & 0.0/ 1.0/ 0.0 & 59.6/ 58.6/ 64.7 & 54.5/ 55.6/ 53.5 \\
\hline
\end{tabular}%
}
\tightfloatafter
\end{table*}

\begin{table*}[htbp]
\tightfloatbefore
\centering
\scriptsize
\setlength{\tabcolsep}{2.8pt}
\renewcommand{\arraystretch}{1.08}
\caption{Detailed RQ1 simulation task progress for stacking bowls. Rows report individual robots, while columns vary the state/action representation. Each entry reports run-1/ run-2/ run-3 task progress (\%).}
\label{tab:supp_rq1_stackbowl}
\resizebox{\textwidth}{!}{%
\begin{tabular}{lcccc}
\hline
\rule{0pt}{2.6ex}\textbf{Target embodiment} & \textbf{Abs. EEF / World-Delta} & \textbf{Abs. EEF / EEF-Delta} & \textbf{EEF-Delta / World-Delta} & \textbf{EEF-Delta / EEF-Delta} \\
\hline
Franka & 82.0/ 90.0/ 87.0 & 95.0/ 93.0/ 89.0 & 84.0/ 84.0/ 88.0 & 91.0/ 87.0/ 86.0 \\
Franka+logo & 87.0/ 83.0/ 83.0 & 95.0/ 87.0/ 89.0 & 87.0/ 86.0/ 89.0 & 79.0/ 88.0/ 87.0 \\
Franka+green & 87.0/ 89.0/ 78.0 & 95.0/ 85.0/ 85.0 & 82.0/ 82.0/ 84.0 & 87.0/ 89.0/ 90.0 \\
Franka+UMI & 77.0/ 83.0/ 77.0 & 86.0/ 86.0/ 81.0 & 82.0/ 73.0/ 75.0 & 73.0/ 77.0/ 69.0 \\
UR5e+Franka & 78.0/ 75.0/ 77.0 & 71.0/ 79.0/ 73.0 & 72.0/ 68.0/ 65.0 & 83.0/ 83.0/ 85.0 \\
Google+Franka & 0.0/ 0.0/ 2.0 & 0.0/ 0.0/ 0.0 & 64.0/ 66.0/ 58.0 & 77.0/ 70.0/ 73.0 \\
UR5e+UMI & 70.0/ 64.0/ 61.0 & 77.0/ 65.0/ 71.0 & 62.0/ 53.0/ 50.0 & 74.0/ 74.0/ 69.0 \\
GoogleRobot & 0.0/ 1.0/ 0.0 & 0.0/ 0.0/ 0.0 & 41.0/ 55.0/ 48.0 & 29.0/ 54.0/ 54.0 \\
\hline
\end{tabular}%
}
\tightfloatafter
\end{table*}

\begin{table*}[htbp]
\tightfloatbefore
\centering
\scriptsize
\setlength{\tabcolsep}{2.8pt}
\renewcommand{\arraystretch}{1.08}
\caption{Detailed RQ1 simulation task progress for stacking cubes. Rows report individual robots, while columns vary the state/action representation. Each entry reports run-1/ run-2/ run-3 task progress (\%).}
\label{tab:supp_rq1_stackcube}
\resizebox{\textwidth}{!}{%
\begin{tabular}{lcccc}
\hline
\rule{0pt}{2.6ex}\textbf{Target embodiment} & \textbf{Abs. EEF / World-Delta} & \textbf{Abs. EEF / EEF-Delta} & \textbf{EEF-Delta / World-Delta} & \textbf{EEF-Delta / EEF-Delta} \\
\hline
Franka & 94.0/ 98.0/ 94.0 & 97.0/ 99.0/ 98.0 & 97.0/ 93.0/ 91.0 & 98.0/ 98.0/ 96.0 \\
Franka+logo & 92.0/ 97.0/ 90.0 & 99.0/ 98.0/ 98.0 & 95.0/ 93.0/ 93.0 & 96.0/ 98.0/ 96.0 \\
Franka+green & 87.0/ 92.0/ 90.0 & 96.0/ 92.0/ 98.0 & 97.0/ 91.0/ 93.0 & 94.0/ 93.0/ 96.0 \\
Franka+UMI & 95.0/ 89.0/ 86.0 & 93.0/ 98.0/ 91.0 & 82.0/ 79.0/ 75.7 & 77.0/ 81.0/ 84.0 \\
UR5e+Franka & 79.0/ 82.0/ 81.0 & 89.0/ 92.0/ 91.0 & 75.0/ 88.0/ 79.0 & 71.0/ 77.0/ 84.0 \\
Google+Franka & 2.0/ 10.0/ 2.0 & 3.0/ 2.0/ 3.0 & 79.0/ 74.0/ 74.0 & 79.0/ 75.0/ 78.0 \\
UR5e+UMI & 79.0/ 71.0/ 67.0 & 84.0/ 80.0/ 85.0 & 59.0/ 58.0/ 61.0 & 69.0/ 63.0/ 66.0 \\
GoogleRobot & 2.0/ 2.0/ 2.0 & 6.0/ 2.0/ 2.0 & 70.0/ 57.0/ 63.0 & 59.0/ 47.0/ 56.0 \\
\hline
\end{tabular}%
}
\tightfloatafter
\end{table*}

\begin{table*}[htbp]
\tightfloatbefore
\centering
\scriptsize
\setlength{\tabcolsep}{2.8pt}
\renewcommand{\arraystretch}{1.08}
\caption{Detailed RQ2 simulation task progress for serving sausages. Rows report individual robots, while columns vary the number of non-target source embodiments in the fixed-budget pretraining pool. Each entry reports run-1/ run-2/ run-3 task progress (\%).}
\label{tab:supp_rq2_sausage}
\resizebox{0.65\textwidth}{!}{%
\begin{tabular}{lccc}
\hline
\rule{0pt}{2.6ex}\textbf{Target embodiment} & \textbf{1 source} & \textbf{8 sources} & \textbf{512 sources} \\
\hline
Franka & 84.8/ 80.8/ 81.8 & 82.8/ 81.8/ 85.9 & 84.8/ 91.9/ 90.9 \\
Franka+logo & 80.8/ 76.8/ 84.8 & 84.8/ 80.8/ 81.8 & 88.9/ 85.9/ 86.9 \\
Franka+green & 78.8/ 81.8/ 79.8 & 80.8/ 84.8/ 85.9 & 85.9/ 91.9/ 86.9 \\
Franka+UMI & 53.5/ 53.5/ 61.6 & 72.7/ 75.8/ 77.8 & 74.8/ 81.8/ 86.9 \\
UR5e+Franka & 40.4/ 50.5/ 45.5 & 52.5/ 52.5/ 58.6 & 78.8/ 81.8/ 75.8 \\
Google+Franka & 41.4/ 45.5/ 45.5 & 56.6/ 41.4/ 53.5 & 50.5/ 58.6/ 56.6 \\
UR5e+UMI & 40.4/ 33.3/ 40.4 & 63.6/ 54.5/ 56.6 & 69.7/ 69.7/ 69.7 \\
GoogleRobot & 31.3/ 28.3/ 26.3 & 47.5/ 45.5/ 43.4 & 54.5/ 55.6/ 53.5 \\
\hline
\end{tabular}%
}
\tightfloatafter
\end{table*}

\begin{table*}[htbp]
\tightfloatbefore
\centering
\scriptsize
\setlength{\tabcolsep}{2.8pt}
\renewcommand{\arraystretch}{1.08}
\caption{Detailed RQ2 simulation task progress for stacking bowls. Rows report individual robots, while columns vary the number of non-target source embodiments in the fixed-budget pretraining pool. Each entry reports run-1/ run-2/ run-3 task progress (\%).}
\label{tab:supp_rq2_stackbowl}
\resizebox{0.65\textwidth}{!}{%
\begin{tabular}{lccc}
\hline
\rule{0pt}{2.6ex}\textbf{Target embodiment} & \textbf{1 source} & \textbf{8 sources} & \textbf{512 sources} \\
\hline
Franka & 87.0/ 81.0/ 84.0 & 92.0/ 87.0/ 85.0 & 91.0/ 87.0/ 86.0 \\
Franka+logo & 84.0/ 84.0/ 83.0 & 89.0/ 87.0/ 93.0 & 79.0/ 88.0/ 87.0 \\
Franka+green & 85.0/ 85.0/ 86.0 & 86.0/ 86.0/ 86.0 & 87.0/ 89.0/ 90.0 \\
Franka+UMI & 64.0/ 65.0/ 58.0 & 68.0/ 71.0/ 67.0 & 73.0/ 77.0/ 69.0 \\
UR5e+Franka & 57.0/ 60.0/ 56.0 & 76.0/ 73.0/ 76.0 & 83.0/ 83.0/ 85.0 \\
Google+Franka & 55.0/ 64.0/ 61.0 & 70.0/ 71.0/ 62.0 & 77.0/ 70.0/ 73.0 \\
UR5e+UMI & 51.0/ 49.0/ 39.0 & 63.0/ 66.0/ 64.0 & 74.0/ 74.0/ 69.0 \\
GoogleRobot & 15.0/ 28.0/ 29.0 & 45.0/ 35.0/ 49.0 & 29.0/ 54.0/ 54.0 \\
\hline
\end{tabular}%
}
\tightfloatafter
\end{table*}

\begin{table*}[htbp]
\tightfloatbefore
\centering
\scriptsize
\setlength{\tabcolsep}{2.8pt}
\renewcommand{\arraystretch}{1.08}
\caption{Detailed RQ2 simulation task progress for stacking cubes. Rows report individual robots, while columns vary the number of non-target source embodiments in the fixed-budget pretraining pool. Each entry reports run-1/ run-2/ run-3 task progress (\%).}
\label{tab:supp_rq2_stackcube}
\resizebox{0.65\textwidth}{!}{%
\begin{tabular}{lccc}
\hline
\rule{0pt}{2.6ex}\textbf{Target embodiment} & \textbf{1 source} & \textbf{8 sources} & \textbf{512 sources} \\
\hline
Franka & 93.0/ 92.0/ 91.0 & 95.0/ 95.0/ 96.0 & 98.0/ 98.0/ 96.0 \\
Franka+logo & 89.0/ 87.0/ 93.0 & 94.0/ 98.0/ 95.0 & 96.0/ 98.0/ 96.0 \\
Franka+green & 91.0/ 89.0/ 87.0 & 96.0/ 97.0/ 98.0 & 94.0/ 93.0/ 96.0 \\
Franka+UMI & 53.0/ 51.0/ 59.0 & 86.0/ 81.0/ 87.0 & 77.0/ 81.0/ 84.0 \\
UR5e+Franka & 55.0/ 59.0/ 60.0 & 68.0/ 69.0/ 63.0 & 71.0/ 77.0/ 84.0 \\
Google+Franka & 60.0/ 64.0/ 59.0 & 66.0/ 69.0/ 67.0 & 79.0/ 75.0/ 78.0 \\
UR5e+UMI & 39.0/ 36.0/ 33.0 & 61.0/ 55.0/ 55.0 & 69.0/ 63.0/ 66.0 \\
GoogleRobot & 39.0/ 35.0/ 28.0 & 43.0/ 46.0/ 52.0 & 59.0/ 47.0/ 56.0 \\
\hline
\end{tabular}%
}
\tightfloatafter
\end{table*}

\begin{table*}[htbp]
\tightfloatbefore
\centering
\scriptsize
\setlength{\tabcolsep}{2.8pt}
\renewcommand{\arraystretch}{1.08}
\caption{Detailed RQ3 simulation task progress for serving sausages. Rows report individual robots, while columns vary the auxiliary co-training objective. Each entry reports run-1/ run-2/ run-3 task progress (\%).}
\label{tab:supp_rq3_sausage}
\resizebox{\textwidth}{!}{%
\begin{tabular}{lccccc}
\hline
\rule{0pt}{2.6ex}\textbf{Target embodiment} & \textbf{No co-training} & \textbf{LAP (EEF frame)} & \textbf{LAP (both frames)} & \textbf{Subgoal} & \textbf{BBox} \\
\hline
Franka & 84.8/ 91.9/ 90.9 & 90.9/ 84.8/ 91.9 & 88.9/ 91.9/ 91.9 & 91.9/ 95.0/ 93.9 & 85.9/ 91.9/ 87.9 \\
Franka+logo & 88.9/ 85.9/ 86.9 & 87.9/ 85.9/ 86.9 & 85.9/ 87.9/ 90.9 & 97.0/ 91.9/ 95.0 & 92.9/ 92.9/ 91.9 \\
Franka+green & 85.9/ 91.9/ 86.9 & 80.8/ 87.9/ 80.8 & 87.9/ 86.9/ 89.9 & 92.9/ 96.0/ 95.0 & 88.9/ 88.9/ 88.9 \\
Franka+UMI & 74.8/ 81.8/ 86.9 & 84.8/ 82.8/ 85.9 & 76.8/ 79.8/ 76.8 & 84.8/ 88.9/ 93.9 & 85.9/ 86.9/ 88.9 \\
UR5e+Franka & 78.8/ 81.8/ 75.8 & 78.8/ 84.8/ 81.8 & 72.7/ 76.8/ 68.7 & 89.9/ 91.9/ 92.9 & 84.8/ 83.8/ 83.8 \\
Google+Franka & 50.5/ 58.6/ 56.6 & 77.8/ 74.8/ 73.7 & 58.6/ 58.6/ 61.6 & 68.7/ 69.7/ 70.7 & 71.7/ 72.7/ 64.7 \\
UR5e+UMI & 69.7/ 69.7/ 69.7 & 77.8/ 72.7/ 74.8 & 70.7/ 74.8/ 70.7 & 80.8/ 78.8/ 79.8 & 79.8/ 85.9/ 73.7 \\
GoogleRobot & 54.5/ 55.6/ 53.5 & 67.7/ 61.6/ 68.7 & 62.6/ 50.5/ 55.6 & 62.6/ 62.6/ 64.7 & 59.6/ 69.7/ 57.6 \\
\hline
\end{tabular}%
}
\tightfloatafter
\end{table*}

\begin{table*}[htbp]
\tightfloatbefore
\centering
\scriptsize
\setlength{\tabcolsep}{2.8pt}
\renewcommand{\arraystretch}{1.08}
\caption{Detailed RQ3 simulation task progress for stacking bowls. Rows report individual robots, while columns vary the auxiliary co-training objective. Each entry reports run-1/ run-2/ run-3 task progress (\%).}
\label{tab:supp_rq3_stackbowl}
\resizebox{\textwidth}{!}{%
\begin{tabular}{lccccc}
\hline
\rule{0pt}{2.6ex}\textbf{Target embodiment} & \textbf{No co-training} & \textbf{LAP (EEF frame)} & \textbf{LAP (both frames)} & \textbf{Subgoal} & \textbf{BBox} \\
\hline
Franka & 91.0/ 87.0/ 86.0 & 84.0/ 87.0/ 89.0 & 81.0/ 86.0/ 87.0 & 80.0/ 82.0/ 86.0 & 87.0/ 90.0/ 86.0 \\
Franka+logo & 79.0/ 88.0/ 87.0 & 85.0/ 89.0/ 84.0 & 92.0/ 89.0/ 84.0 & 85.0/ 79.0/ 84.0 & 81.0/ 89.0/ 85.0 \\
Franka+green & 87.0/ 89.0/ 90.0 & 88.0/ 85.0/ 85.0 & 82.0/ 82.0/ 83.0 & 85.0/ 76.0/ 84.0 & 89.0/ 83.0/ 86.0 \\
Franka+UMI & 73.0/ 77.0/ 69.0 & 86.0/ 80.0/ 82.0 & 74.0/ 74.0/ 75.0 & 78.0/ 82.0/ 86.0 & 79.0/ 79.0/ 82.0 \\
UR5e+Franka & 83.0/ 83.0/ 85.0 & 83.0/ 86.0/ 80.0 & 73.0/ 84.0/ 82.0 & 83.0/ 80.0/ 86.0 & 79.0/ 76.0/ 79.0 \\
Google+Franka & 77.0/ 70.0/ 73.0 & 69.0/ 68.0/ 69.0 & 76.0/ 72.0/ 61.0 & 73.0/ 68.0/ 60.0 & 76.0/ 83.0/ 74.0 \\
UR5e+UMI & 74.0/ 74.0/ 69.0 & 73.0/ 80.0/ 71.0 & 70.0/ 65.0/ 68.0 & 83.0/ 72.0/ 81.0 & 74.0/ 72.0/ 80.0 \\
GoogleRobot & 29.0/ 54.0/ 54.0 & 65.0/ 68.0/ 58.0 & 62.0/ 55.0/ 65.0 & 31.0/ 55.0/ 57.0 & 64.0/ 67.0/ 64.0 \\
\hline
\end{tabular}%
}
\tightfloatafter
\end{table*}

\begin{table*}[htbp]
\tightfloatbefore
\centering
\scriptsize
\setlength{\tabcolsep}{2.8pt}
\renewcommand{\arraystretch}{1.08}
\caption{Detailed RQ3 simulation task progress for stacking cubes. Rows report individual robots, while columns vary the auxiliary co-training objective. Each entry reports run-1/ run-2/ run-3 task progress (\%).}
\label{tab:supp_rq3_stackcube}
\resizebox{\textwidth}{!}{%
\begin{tabular}{lccccc}
\hline
\rule{0pt}{2.6ex}\textbf{Target embodiment} & \textbf{No co-training} & \textbf{LAP (EEF frame)} & \textbf{LAP (both frames)} & \textbf{Subgoal} & \textbf{BBox} \\
\hline
Franka & 98.0/ 98.0/ 96.0 & 96.0/ 96.0/ 95.0 & 93.0/ 91.0/ 95.0 & 89.0/ 90.0/ 94.0 & 96.0/ 95.0/ 94.0 \\
Franka+logo & 96.0/ 98.0/ 96.0 & 94.0/ 94.0/ 96.0 & 91.0/ 94.0/ 95.0 & 94.0/ 94.0/ 93.0 & 96.0/ 95.0/ 94.0 \\
Franka+green & 94.0/ 93.0/ 96.0 & 97.0/ 91.0/ 97.0 & 92.0/ 89.0/ 92.0 & 92.0/ 98.0/ 94.0 & 96.0/ 94.0/ 92.0 \\
Franka+UMI & 77.0/ 81.0/ 84.0 & 95.0/ 93.0/ 91.0 & 86.0/ 85.0/ 91.0 & 91.0/ 92.0/ 90.0 & 94.0/ 93.0/ 94.0 \\
UR5e+Franka & 71.0/ 77.0/ 84.0 & 92.0/ 90.0/ 88.0 & 86.0/ 88.0/ 83.0 & 80.0/ 83.0/ 83.0 & 91.0/ 84.0/ 89.0 \\
Google+Franka & 79.0/ 75.0/ 78.0 & 74.0/ 79.0/ 77.0 & 68.0/ 77.0/ 82.0 & 73.0/ 77.0/ 70.0 & 84.0/ 80.0/ 85.0 \\
UR5e+UMI & 69.0/ 63.0/ 66.0 & 86.0/ 84.0/ 86.0 & 87.0/ 80.0/ 86.0 & 86.0/ 89.0/ 87.0 & 83.0/ 89.0/ 87.0 \\
GoogleRobot & 59.0/ 47.0/ 56.0 & 56.0/ 58.0/ 60.0 & 66.0/ 73.0/ 70.0 & 75.0/ 71.0/ 71.0 & 61.0/ 59.0/ 68.0 \\
\hline
\end{tabular}%
}
\tightfloatafter
\end{table*}

\begin{table*}[htbp]
\tightfloatbefore
\centering
\scriptsize
\setlength{\tabcolsep}{2.8pt}
\renewcommand{\arraystretch}{1.08}
\caption{Detailed RQ4 simulation task progress for serving sausages. Rows vary the target-embodiment pretraining exposure ratio and the target post-training oracle; columns report the two target embodiments used in the sweep. Each entry reports run-1/ run-2/ run-3 task progress (\%).}
\label{tab:supp_rq4_sausage}
\resizebox{0.6\textwidth}{!}{%
\begin{tabular}{lcc}
\hline
\rule{0pt}{2.6ex}\textbf{Setting} & \textbf{UR5e+UMI} & \textbf{GoogleRobot} \\
\hline
Strict zero-shot (0\%) & 69.7/ 69.7/ 69.7 & 54.5/ 55.6/ 53.5 \\
Pretrain-exposed (5\%) & 75.8/ 73.7/ 74.8 & 68.7/ 64.7/ 68.7 \\
Pretrain-exposed (30\%) & 86.9/ 83.8/ 75.8 & 71.7/ 72.7/ 67.7 \\
Pretrain-exposed (100\%) & 89.9/ 84.8/ 85.9 & 73.7/ 68.7/ 71.7 \\
Target post-training oracle & 83.8/ 78.8/ 80.8 & 69.7/ 62.6/ 60.6 \\
\hline
\end{tabular}%
}
\tightfloatafter
\end{table*}

\begin{table*}[htbp]
\tightfloatbefore
\centering
\scriptsize
\setlength{\tabcolsep}{2.8pt}
\renewcommand{\arraystretch}{1.08}
\caption{Detailed RQ4 simulation task progress for stacking bowls. Rows vary the target-embodiment pretraining exposure ratio and the target post-training oracle; columns report the two target embodiments used in the sweep. Each entry reports run-1/ run-2/ run-3 task progress (\%).}
\label{tab:supp_rq4_stackbowl}
\resizebox{0.6\textwidth}{!}{%
\begin{tabular}{lcc}
\hline
\rule{0pt}{2.6ex}\textbf{Setting} & \textbf{UR5e+UMI} & \textbf{GoogleRobot} \\
\hline
Strict zero-shot (0\%) & 74.0/ 74.0/ 69.0 & 29.0/ 54.0/ 54.0 \\
Pretrain-exposed (5\%) & 80.0/ 77.0/ 80.0 & 70.0/ 61.0/ 67.0 \\
Pretrain-exposed (30\%) & 87.0/ 82.0/ 85.0 & 70.0/ 69.0/ 59.0 \\
Pretrain-exposed (100\%) & 88.0/ 81.0/ 86.0 & 71.0/ 61.0/ 71.0 \\
Target post-training oracle & 97.0/ 92.0/ 92.0 & 81.0/ 79.0/ 76.0 \\
\hline
\end{tabular}%
}
\tightfloatafter
\end{table*}

\begin{table*}[htbp]
\tightfloatbefore
\centering
\scriptsize
\setlength{\tabcolsep}{2.8pt}
\renewcommand{\arraystretch}{1.08}
\caption{Detailed RQ4 simulation task progress for stacking cubes. Rows vary the target-embodiment pretraining exposure ratio and the target post-training oracle; columns report the two target embodiments used in the sweep. Each entry reports run-1/ run-2/ run-3 task progress (\%).}
\label{tab:supp_rq4_stackcube}
\resizebox{0.6\textwidth}{!}{%
\begin{tabular}{lcc}
\hline
\rule{0pt}{2.6ex}\textbf{Setting} & \textbf{UR5e+UMI} & \textbf{GoogleRobot} \\
\hline
Strict zero-shot (0\%) & 69.0/ 63.0/ 66.0 & 59.0/ 47.0/ 56.0 \\
Pretrain-exposed (5\%) & 87.0/ 79.0/ 80.0 & 75.0/ 68.0/ 77.0 \\
Pretrain-exposed (30\%) & 91.0/ 90.0/ 87.0 & 76.0/ 68.0/ 67.0 \\
Pretrain-exposed (100\%) & 92.0/ 84.0/ 84.0 & 77.0/ 71.0/ 68.0 \\
Target post-training oracle & 98.0/ 95.0/ 93.0 & 90.0/ 88.0/ 89.0 \\
\hline
\end{tabular}%
}
\tightfloatafter
\end{table*}

\begin{table*}[htbp]
\tightfloatbefore
\centering
\scriptsize
\setlength{\tabcolsep}{2.8pt}
\renewcommand{\arraystretch}{1.08}
\caption{Detailed RQ1 real-world task progress for \emph{open the fryer}. Rows report individual robots, while columns vary the state/action representation. Each entry reports mean task progress (\%) over 10 rollouts.}
\label{tab:supp_rq1_real_fryer}
\resizebox{\textwidth}{!}{%
\begin{tabular}{lcccc}
\hline
\rule{0pt}{2.6ex}\textbf{Target embodiment} & \textbf{Abs. EEF / World-Delta} & \textbf{Abs. EEF / EEF-Delta} & \textbf{EEF-Delta / World-Delta} & \textbf{EEF-Delta / EEF-Delta} \\
\hline
Franka+Robotiq & 90.0 & 95.0 & 80.0 & 100.0 \\
Franka+Robotiq (CoRL logo) & 90.0 & 90.0 & 90.0 & 100.0 \\
Franka+Umi & 80.0 & 85.0 & 90.0 & 100.0 \\
UR & 55.0 & 35.0 & 55.0 & 90.0 \\
Humanoid robot & 0.0 & 0.0 & 35.0 & 100.0 \\
UR+Umi & 60.0 & 50.0 & 60.0 & 80.0 \\
Humanoid robot+Umi & 0.0 & 0.0 & 55.0 & 85.0 \\
Piper-on-Quadruped & 35.0 & 0.0 & 40.0 & 80.0 \\
\hline
\end{tabular}%
}
\tightfloatafter
\end{table*}

\begin{table*}[htbp]
\tightfloatbefore
\centering
\scriptsize
\setlength{\tabcolsep}{2.8pt}
\renewcommand{\arraystretch}{1.08}
\caption{Detailed RQ1 real-world task progress for \emph{water the flower}. Rows report individual robots, while columns vary the state/action representation. Each entry reports mean task progress (\%) over 10 rollouts.}
\label{tab:supp_rq1_real_flower}
\resizebox{\textwidth}{!}{%
\begin{tabular}{lcccc}
\hline
\rule{0pt}{2.6ex}\textbf{Target embodiment} & \textbf{Abs. EEF / World-Delta} & \textbf{Abs. EEF / EEF-Delta} & \textbf{EEF-Delta / World-Delta} & \textbf{EEF-Delta / EEF-Delta} \\
\hline
Franka+Robotiq & 85.0 & 85.0 & 90.0 & 100.0 \\
Franka+Robotiq (CoRL logo) & 85.0 & 90.0 & 65.0 & 95.0 \\
Franka+Umi & 55.0 & 75.0 & 30.0 & 85.0 \\
UR & 40.0 & 85.0 & 70.0 & 85.0 \\
Humanoid robot & 0.0 & 0.0 & 40.0 & 80.0 \\
UR+Umi & 55.0 & 60.0 & 60.0 & 75.0 \\
Humanoid robot+Umi & 0.0 & 0.0 & 25.0 & 65.0 \\
Piper-on-Quadruped & 20.0 & 20.0 & 30.0 & 90.0 \\
\hline
\end{tabular}%
}
\tightfloatafter
\end{table*}

\begin{table*}[htbp]
\tightfloatbefore
\centering
\scriptsize
\setlength{\tabcolsep}{4pt}
\renewcommand{\arraystretch}{1.08}
\caption{Detailed RQ2 real-world task progress for \emph{open the fryer}. Rows report individual robots, while columns vary the number of pretraining embodiments $|\mathcal{E}_{\mathrm{pre}}|$. Each entry reports mean task progress (\%) over 10 rollouts.}
\label{tab:supp_rq2_real_fryer}
\resizebox{0.8\textwidth}{!}{%
\begin{tabular}{lccc}
\hline
\rule{0pt}{2.6ex}\textbf{Target embodiment} & \textbf{1 source} & \textbf{8 sources} & \textbf{512 sources} \\
\hline
Franka+Robotiq & 80.0 & 100.0 & 100.0 \\
Franka+Robotiq (CoRL logo) & 90.0 & 100.0 & 100.0 \\
Franka+Umi & 80.0 & 100.0 & 100.0 \\
UR & 75.0 & 75.0 & 90.0 \\
Humanoid robot & 95.0 & 90.0 & 100.0 \\
UR+Umi & 65.0 & 75.0 & 80.0 \\
Humanoid robot+Umi & 75.0 & 70.0 & 85.0 \\
Piper-on-Quadruped & 90.0 & 45.0 & 80.0 \\
\hline
\end{tabular}%
}
\tightfloatafter
\end{table*}

\begin{table*}[htbp]
\tightfloatbefore
\centering
\scriptsize
\setlength{\tabcolsep}{4pt}
\renewcommand{\arraystretch}{1.08}
\caption{Detailed RQ2 real-world task progress for \emph{water the flower}. Rows report individual robots, while columns vary the number of pretraining embodiments $|\mathcal{E}_{\mathrm{pre}}|$. Each entry reports mean task progress (\%) over 10 rollouts.}
\label{tab:supp_rq2_real_flower}
\resizebox{0.8\textwidth}{!}{%
\begin{tabular}{lccc}
\hline
\rule{0pt}{2.6ex}\textbf{Target embodiment} & \textbf{1 source} & \textbf{8 sources} & \textbf{512 sources} \\
\hline
Franka+Robotiq & 80.0 & 95.0 & 100.0 \\
Franka+Robotiq (CoRL logo) & 85.0 & 85.0 & 95.0 \\
Franka+Umi & 30.0 & 65.0 & 85.0 \\
UR & 75.0 & 35.0 & 85.0 \\
Humanoid robot & 85.0 & 30.0 & 80.0 \\
UR+Umi & 55.0 & 35.0 & 75.0 \\
Humanoid robot+Umi & 50.0 & 30.0 & 65.0 \\
Piper-on-Quadruped & 65.0 & 70.0 & 90.0 \\
\hline
\end{tabular}%
}
\tightfloatafter
\end{table*}

The simulation tables in Tables~\ref{tab:supp_rq1_sausage}--\ref{tab:supp_rq4_stackcube} expand the aggregate results in the main paper into per-task, per-embodiment progress scores.
Unless otherwise stated, each entry reports the three simulation repeats as run-1/ run-2/ run-3 progress.
We report the source embodiment and the seven held-out target embodiments used in the main simulation study.
For real-world evaluation, Tables~\ref{tab:supp_rq1_real_fryer}--\ref{tab:supp_rq2_real_flower} report the completed RQ1 and RQ2 runs on \emph{open the fryer} and \emph{water the flower}. Each task is reported in a separate table, target embodiments are rows, and the controlled experimental variable is shown in columns.

\tightsection{Taxonomy of Zero-Shot Protocols in Prior Work}
\label{sec:zeroshot_taxonomy}
We revisit representative prior cross-embodiment studies and categorize their evaluation protocols based on target-embodiment exposure during pretraining and post-training (Table~\ref{tab:zeroshot_taxonomy}).
This analysis reveals inconsistent uses of ``zero-shot'' and ``out-of-the-box'' transfer across prior work, with some evaluations involving unseen task--embodiment pairs rather than entirely unseen target embodiments.
Such protocol differences complicate comparisons of reported transfer performance, motivating our explicit distinction between strict and pretrain-exposed zero-shot transfer and the corresponding controlled evaluations.
\begin{table*}[h]
    \tightfloatbefore
    \centering
    \caption{
        Classification of representative prior cross-embodiment manipulation evaluations described as zero-shot or out-of-the-box in the literature,
        reclassified according to target-embodiment exposure under our protocol taxonomy.
    }
    \label{tab:zeroshot_taxonomy}
    \footnotesize
    \setlength{\tabcolsep}{3pt}
    \renewcommand{\arraystretch}{0.92}
    \setlength{\aboverulesep}{0.3ex}
    \setlength{\belowrulesep}{0.3ex}
    \begin{tabular}{@{}lccc@{}}
    \toprule
    \textbf{Work} &
    \textbf{Target embodiment} &
    \textbf{Target embodiment} &
    \textbf{Protocol} \\
    &
    \textbf{in pretraining?} &
    \textbf{in post-training?} & \\
    \midrule

    LAP~\cite{lap}
    & \ding{55} & \ding{55} & \textbf{Strict} \\

    Cloak~\cite{piseno2026cloak}
    & \ding{55} & \ding{55} & \textbf{Strict} \\

    RDT2~\cite{liu2026rdt2}
    & \ding{55} & \ding{55} & \textbf{Strict} \\

    Octo~\cite{octo2024}
    & \ding{51} & \ding{55} & \textbf{Pretrain-exposed} \\

    $\pi_{0.7}$~\cite{intelligence2026pi07}
    & \ding{51} & \ding{55} & \textbf{Pretrain-exposed} \\

    Qwen-RobotManip~\cite{yuan2026qwenrobotmanip}
    & \ding{51} & \ding{55} & \textbf{Pretrain-exposed} \\

    OpenVLA~\cite{kim2024openvla}
    & \ding{51} & \ding{55} & \textbf{Pretrain-exposed} \\

    Gemini Robotics 1.5~\cite{geminiroboticsteam2025geminirobotics15}$^\ast$
    & \ding{51} & \ding{55} & \textbf{Pretrain-exposed} \\

    Being-H0.5~\cite{luo2026being}$^\dagger$
    & \ding{51} & \ding{51} & \textbf{Unseen task--emb. pair} \\

    \bottomrule
    \end{tabular}\\[0.25em]
    \noindent\begin{minipage}{\textwidth}
    {\scriptsize\sloppy
    $^\ast$Gemini Robotics 1.5 reports zero-shot transfer to tasks observed only on another embodiment, while the evaluated target embodiments are already included in its multi-embodiment pretraining and receive no robot-specific post-training.

    $^\dagger$Being-H0.5 reports unseen task--embodiment pairs as zero-shot even though the target embodiment participates in post-training, so the evaluation falls outside both strict and pretrain-exposed protocols.\par}
    \end{minipage}
    \tightfloatafter
\end{table*}

\tightsection{Cross-Embodiment Feature Visualization}
\label{sec:rq2_umap}

\begin{wrapfigure}{r}{0.38\columnwidth}
\vspace{-0.7cm}
\centering
\includegraphics[width=\linewidth,trim=0 0 52bp 0,clip]{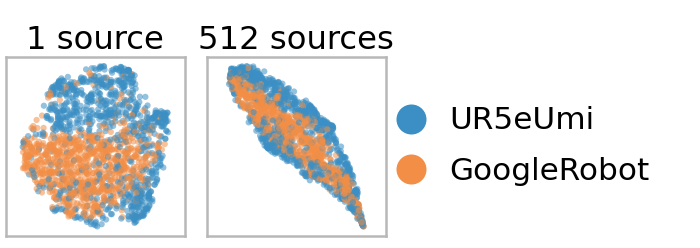}
\caption{UMAP of mean-pooled last-layer VLM features from held-out UR5eUMI and GoogleRobot observations.}
\label{fig:rq2_feature_umap}
\vspace{-0.3cm}
\end{wrapfigure}

To complement the quantitative results in RQ2, we visualize mean-pooled last-layer VLM features using UMAP with cosine distance. We sample 3,600 frames from 1,200 trajectories across two unseen target embodiments, UR5eUMI and GoogleRobot.
As shown in Fig.~\ref{fig:rq2_feature_umap}, the 512-source model exhibits qualitatively greater feature overlap between the two embodiments than the single-source model, providing preliminary evidence of stronger cross-embodiment feature alignment.

\tightsection{Data-Construction Details}
\label{sec:supp_data_construction}

\begin{figure}[htbp]
    \tightfloatbefore
    \centering
    \includegraphics[width=\linewidth]{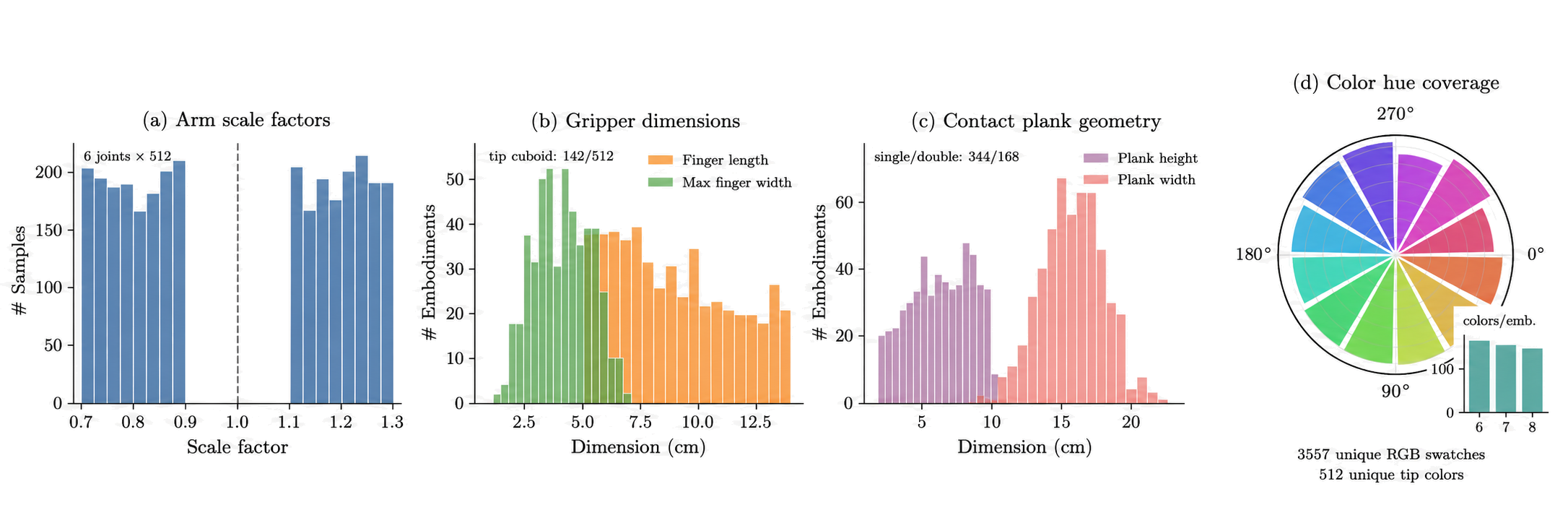}
    \caption{Empirical distributions of the procedurally generated source-embodiment pool.
    We summarize the arm-scale factors, sampled gripper dimensions, contact-plank geometry, and color-hue coverage for the 512 generated Franka-style tabletop embodiments used for source pretraining.}
    \label{fig:supp_embodiment_generation_histograms}
    \tightfloatafter
\end{figure}

\textbf{Simulation embodiment generation.}
The 512 source embodiments are procedurally generated from a Franka-style tabletop arm template. We introduce embodiment diversity along geometry, end-effector morphology, and visual appearance.
For arm-level variation, we scale the six movable arm-link regions with factors sampled from shrink and expansion modes, spanning approximately $0.7$--$1.3$, and consistently deform the corresponding visual and collision meshes.
For end-effector variation, we procedurally generate gripper fingers by sampling finger length, width and thickness profiles, gripper rotation, and optional fingertip cuboids.
We also randomize the contact plank attached to the gripper, including its height, width, single- or double-layer structure, and upper-surface shape.
For visual variation, each embodiment is assigned a balanced low-saturation multi-color palette with 6--8 colors, which is applied to the arm, finger, and plank meshes; fingertip colors are also varied across embodiments.
\autoref{fig:supp_embodiment_generation_histograms} reports post-generation statistics with histograms of arm scale factors, gripper dimensions, contact-plank geometry, and color-hue coverage.
These statistics show that the generated embodiments span diverse arm geometries, gripper shapes, contact surfaces, and appearances while sharing a common Franka-style tabletop manipulation setup.
The seven commercially used test robots are imported as external held-out targets and are excluded from this procedurally generated source pool.

\begin{figure}[htbp]
    \tightfloatbefore
    \centering
    \includegraphics[width=0.78\linewidth]{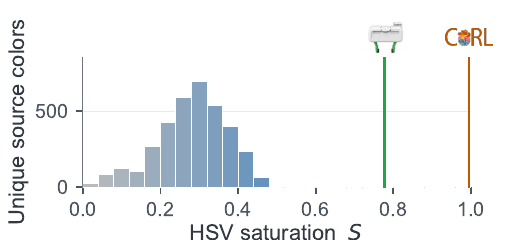}
    \caption{HSV color distribution of the procedurally generated source palette versus the appearance-only evaluation targets.
    The held-out green and logo-orange colors lie outside the source palette primarily along the saturation dimension.}
    \label{fig:appearance_color_distribution}
    \vspace{0.2cm}
    \tightfloatafter
\end{figure}

The appearance-only evaluation targets are near the source visual distribution but are not directly included in the pretraining pool.
In HSV space, the distinguishing green and logo-orange colors have saturation values of 0.77 and 0.99, respectively, compared with a maximum saturation of 0.46 in the generated source palette.
Thus, these targets are held out primarily along the saturation dimension, as shown in \autoref{fig:appearance_color_distribution}.
Under our exposure-based definition, they satisfy strict zero-shot transfer because the target appearances are absent from all training data, although they should not be interpreted as strongly out-of-distribution appearance shifts.

\textbf{Domain randomization.}
Inspired by prior sim-to-real studies~\cite{deng2025graspvlagraspingfoundationmodel, interndataa1, yin2026geniesim30, maddukuri2025simandrealcotrainingsimplerecipe}, we apply extensive domain randomization during pretraining to reduce the sim-to-real gap and prevent the policy from overfitting to simulator-specific visual or physical regularities.
For visual appearance, we use more than 1,000 materials to randomize the background and tabletop textures.
We also randomize illumination intensity, color, type, the number of lights, and light positions, covering a broad range of lighting conditions that can arise in real robot workspaces.
For objects and scene layouts, we filter Objaverse to obtain more than 10,000 manipulation objects for pretraining.
We further randomize tabletop height, size, and pose, and randomly drop distractor objects into the scene to approximate cluttered real-world setups.
For camera configuration, we randomize the extrinsics of both third-person and wrist cameras.
To emulate the diverse camera placements found in real datasets, the third-person camera is randomly rotated around the tabletop center by $[-150^\circ,150^\circ]$, with a pitch angle sampled from $[15^\circ,45^\circ]$ and a distance to the tabletop center sampled from $[0.8,1.4]$~m.
For the wrist view, we randomize the camera extrinsics within a $0.1$~m cube along the gripper direction while rejecting configurations in which the view is occluded by the robot arm itself.
Finally, we randomize physical parameters such as object mass and friction coefficients.
In contrast, post-training data are constructed to mimic small-scale real-world data collection: for each downstream setting, we keep a single scene, object combination, lighting condition, and camera placement fixed, and evaluate the policy under the corresponding matched condition.
% Following prior sim-to-real practice\addcitation, we randomize physical and visual properties under simple quasi-static assumptions to reduce the chance that simulation-specific appearance or contact parameters dominate the measured embodiment-transfer effects.

\tightsection{Embodiment Shift Taxonomy}
\label{sec:supp_embodiment_shift_taxonomy}

\tightsubsection{Simulation}
\label{sec:supp_embodiment_shift_taxonomy_sim}

\textbf{Franka.}
This is the simulation source embodiment, consisting of a Franka-style 7-DoF arm and a Franka two-finger hand; it defines the reference morphology, kinematics, and end-effector interface for the simulated taxonomy.

\textbf{Franka with CoRL logo.}
This target is an appearance-only shift: it uses the same Franka arm and Franka hand as the source, but changes the robot texture by adding the CoRL logo.

\textbf{Franka with green fingers.}
This target is also an appearance-only shift: the arm, hand geometry, and kinematics remain those of the source Franka, while only the visual appearance of the fingers is changed.

\textbf{Franka with UMI gripper.}
This target is a gripper-only shift: it keeps the Franka arm fixed and replaces the Franka hand with a UMI-style two-finger gripper, isolating the effect of end-effector geometry.

\textbf{UR5e with Franka hand.}
This target is an arm-only shift: it replaces the Franka arm with a 6-DoF UR5e arm while retaining a Franka-style two-finger hand, changing arm morphology and kinematics without changing the end effector.

\textbf{GoogleRobot with Franka hand.}
This target is an arm-only shift: it uses the GoogleRobot mobile-manipulator platform with its 7-DoF arm and a Franka-style two-finger hand, changing the manipulator morphology while preserving the evaluated end-effector geometry.
Although GoogleRobot has a mobile base, base states and commands are excluded from the model inputs and outputs to keep the evaluated control interface consistent.

\textbf{UR5e with UMI gripper.}
This target is a full-embodiment shift: both the arm and the end effector differ from the source, combining a 6-DoF UR5e arm with a UMI-style two-finger gripper.

\textbf{GoogleRobot.}
This target is a full-embodiment shift: it uses the GoogleRobot mobile-manipulator platform with a 7-DoF arm and native two-finger gripper, thereby changing both manipulator morphology and end-effector geometry.
As above, mobile-base states and commands are not included in the model inputs or outputs.

\tightsubsection{Real World}
\label{sec:supp_embodiment_shift_taxonomy_real}

\textbf{Franka with Robotiq gripper.}
This is the real-world source embodiment, composed of a Franka arm and a Robotiq two-finger gripper.

\textbf{Franka with CoRL logo and Robotiq gripper.}
This target is an appearance-only shift: it keeps the Franka arm and Robotiq gripper unchanged while modifying only the visible robot appearance with a CoRL logo.

\textbf{Franka with UMI gripper.}
This target is a gripper-only shift: it keeps the Franka arm fixed and replaces the Robotiq gripper with a UMI-style two-finger gripper.

\textbf{UR5e with Robotiq gripper.}
This target is an arm-only shift: it replaces the Franka arm with a 6-DoF UR5e arm while retaining the Robotiq two-finger gripper, isolating the change in arm morphology and kinematics.

\textbf{Humanoid robot.}
This target is a full-embodiment shift: it uses a mobile humanoid platform with a 7-DoF right arm and native two-finger gripper, changing both the manipulator morphology and end-effector geometry relative to the Franka-Robotiq source.
Although the platform is a dual-arm mobile robot, we retain only right-arm information in the model inputs and outputs; the left arm is fixed in a natural lowered pose, and mobile-base states and commands are excluded.

\textbf{UR5e with UMI gripper.}
This target is a full-embodiment shift: it combines a 6-DoF UR5e arm with a UMI-style two-finger gripper, so both the arm and end-effector geometry differ from the source embodiment.

\textbf{Humanoid robot with UMI gripper.}
This target is a full-embodiment shift: it keeps the mobile humanoid platform and 7-DoF right arm while replacing the native gripper with a UMI-style two-finger gripper.
As with the native-gripper humanoid setting, only the right arm is exposed to the policy, the left arm is fixed in a natural lowered pose, and mobile-base states and commands are excluded from the model interface.

\textbf{Piper-on-Quadruped.}
This target is a full-embodiment shift: it mounts a 6-DoF Piper robotic arm with its matched two-finger gripper on the back of a Unitree B2 quadruped platform, introducing a legged mobile base together with a different arm and end effector.
The quadruped base is not represented in the model inputs or outputs, so evaluation remains focused on the manipulator and gripper control interface.

\tightsection{Experimental Setup Details}
\label{sec:supp_experimental_setup_details}

\tightsubsection{Simulation Platform}
We use MuJoCo for physics simulation and IsaacSim for rendering throughout the simulation experiments.

\tightsubsection{Camera Calibration Details}

To reduce embodiment-dependent visual discrepancies, we calibrate camera poses so that, under a canonical robot pose, the image-plane gripper distribution remains close to the visual distribution observed during post-training.
We define a shared tabletop task frame as an environment-anchored coordinate frame fixed to the tabletop workspace, rather than to any particular robot embodiment.
Its horizontal axes lie on the tabletop plane, and its vertical axis is aligned with the table normal.
Each robot expresses object poses, goal poses, and task-relevant spatial quantities in this common frame through a calibrated transformation from its own base frame.
This convention decouples task specification and camera alignment from embodiment-specific base coordinates and kinematic structures.

In simulation, all third-person cameras are strictly aligned with respect to this task frame.
For each evaluation trajectory, the extrinsic pose of the third-person camera is identical across robots when expressed in the corresponding task frame.
This ensures that visual differences across embodiments primarily reflect robot appearance and motion rather than avoidable viewpoint changes.
For wrist-mounted cameras, we apply an analogous alignment rule in the gripper frame: the camera extrinsics are matched across robots when expressed relative to each robot's gripper coordinate frame.
For real-world experiments, we follow the same principle during physical camera setup and calibration, aligning the cameras so that the visible workspace and the gripper's image-plane coverage are approximately consistent across embodiments.

\tightsubsection{Environment and Object Set Details}

To isolate embodiment shift from variation in the workspace, we standardize the evaluation environment as much as possible across robots.
In simulation, all robot embodiments are evaluated under strictly matched scene configurations: the lighting setup, background material, tabletop material, tabletop workspace, and table arrangement are kept identical for the corresponding test condition.
In the real-world experiments, we similarly standardize the physical setup by enclosing the experimental area with black cloth and covering the table with a gray tablecloth, while keeping the table arrangement and workspace layout aligned across robots.
These controls reduce appearance and layout differences that are unrelated to the robot embodiment being evaluated.

We also keep the object set fixed between post-training and evaluation.
For both simulation and real-world experiments, all objects used in the same evaluation trial are identical across robot embodiments.
Moreover, the evaluation objects are exactly the same object instances as those used in the corresponding post-training data, so no additional object-identity shift is introduced at test time.
This design ensures that differences in zero-shot performance primarily reflect embodiment transfer rather than changes in objects, scene layout, or workspace appearance.

\tightsubsection{Details of Metrics}
Each rollout is assigned a task-progress score $s \in \{0, 0.5, 1\}$.
A rollout receives $s=1$ if it fully completes the task, $s=0.5$ if the robot reaches the necessary interaction subgoal by contacting the task-relevant object, and $s=0$ otherwise.
For example, in the serving-sausages task, contacting the sausage receives half credit, while successfully serving the sausage receives full credit.
All reported progress values are expressed as percentages, i.e., $100s$ averaged over rollouts.

We use a macro-averaging protocol so that each task and each embodiment-shift category contributes uniformly to the aggregate score.
For a given training run, model, and robot embodiment, we first average progress over the downstream tasks.
Within each of the four shift categories, we then average these per-robot task means over all target robots in that category.
The overall average is the arithmetic mean of the four category scores.
For simulation experiments with three independent training runs, we compute the same aggregation separately for each run and report the mean over the three run-level aggregates; when an uncertainty term is shown, it is the standard deviation across these three run-level aggregates.
% progress score

\tightsubsection{Downstream Tasks}
For real-world post-training, we use two tasks, opening an air fryer and watering a plant, with 50 demonstrations per task.
For the simulation benchmark, we use three downstream tasks, serving sausages, stacking bowls, and stacking cubes, with 40K trajectories per task.
To ensure fair comparison, downstream post-training data is collected on the same source Franka arm.

\tightsection{Auxiliary Co-Training Objective Specifications}
\label{sec:supp_auxiliary_cotraining_details}

All auxiliary co-training variants use the same visual input as the imitation policy: the third view and wrist view images, the task caption $\ell$, and the current state when required.
For each variant, the data loader samples imitation-action examples and auxiliary-query examples with weights 0.75 and 0.25, respectively.
The imitation branch remains supervised by the flow-matching action loss, while auxiliary responses are supervised with next-token cross entropy.
The construction rules for the auxiliary answer strings are described below, and \autoref{tab:supp_aux_cotrain_specs} summarizes the exact query templates and example response formats used by the official RQ3 models.
% \autoref{tab:supp_aux_cotrain_specs} summarizes the exact query templates and response formats used by the official RQ3 models.

We follow~\cite{lap} to generate the auxiliary language targets for the language-action objectives. Each target is generated from the delta action vector.
In the EEF-frame variant, the action is represented as $[\Delta x,\Delta y,\Delta z,\Delta r,\Delta p,\Delta \psi,g]$, with translations converted to centimeters and rotations to degrees, then rounded to integers.
Nonzero translation components are rendered as language clauses, nonzero rotation components are rendered as rotation clauses, and the gripper command is rendered as an open-or-close clause.
The EEF-frame translation directions are ordered as up/down, right/left, and forward/backward.
The mixed-frame variant uses the same clause-level format, with the prompt specifying whether the delta is interpreted in the end-effector frame or the robot base frame.
For base-frame labels, translation directions are ordered as forward/backward, left/right, and up/down; roll, pitch, and yaw use right/left, back/forward, and counterclockwise/clockwise, respectively.
This variant tests whether mixing frame semantics in the auxiliary language target helps or hurts the EEF-centered policy representation.

For the subgoal-prediction objective, the answer is selected from the next manipulation-relevant waypoint in the trajectory and encoded as a decimal list in the current EEF coordinate frame.
Translation values are formatted to 0.001 precision and rotation values to 0.01 precision.

For the task-conditioned bounding-box objective, the answer contains one object box per input view, ordered by the configured camera list.
Each box is represented in XYXY order, normalized from image pixels to $[-1,1]$, and emitted as quantized coordinate tokens using 256 uniform bins.
The task-conditioned wording requires the model to identify the object relevant to the instruction rather than predict generic object boxes.

\begin{table*}[htbp]
\tightfloatbefore
\centering
\small
\setlength{\tabcolsep}{4pt}
\caption{Auxiliary co-training objective specifications. Braces denote fields filled from the trajectory annotation. The target column lists example answer formats.}
\label{tab:supp_aux_cotrain_specs}
\resizebox{0.99\textwidth}{!}{%
\begin{tabular}{@{}>{\raggedright\arraybackslash}p{0.15\textwidth}
                >{\raggedright\arraybackslash}p{0.35\textwidth}
                >{\raggedright\arraybackslash}p{0.45\textwidth}@{}}
\hline
\rule{0pt}{2.6ex}\textbf{Objective} & \rule{0pt}{2.6ex}\textbf{Prompt template} & \rule{0pt}{2.6ex}\textbf{Target format} \\
\hline
Language-action (EEF frame)
&
\texttt{The task is \{instruction\}. Predict the robot's action in the end-effector frame; State: \{state tokens\}; Answer:}
&
\texttt{move \{direction\} \{value\} cm}; \texttt{tilt/rotate \{direction\} \{value\} degrees}; \texttt{open gripper}; \texttt{close gripper}. \\
\hline
Language-action (mixed frame)
&
\texttt{The task is \{instruction\}. Predict the robot's action in the \{action frame\}; State: \{state tokens\}; Answer:}
&
\texttt{move \{direction\} \{value\} cm}; \texttt{rotate \{direction\} \{value\} degrees}; \texttt{close gripper}. \\
\hline
Subgoal prediction
&
\texttt{In order to \{instruction\}, what is the next goal pose as xyz and rpy? Answer:}
&
\texttt{[x,y,z,roll,pitch,yaw]}, e.g., \texttt{[0.012,-0.004,0.085,1.57,0.00,-0.12]}. \\
\hline
Task-conditioned BBox
&
\texttt{In order to \{instruction\}, what is the bounding box of the object to be manipulated in the images? Answer:}
&
\texttt{[x\_min,y\_min,x\_max,y\_max]} per view, e.g., \texttt{[[q\_xmin,q\_ymin,q\_xmax,q\_ymax], ...]}. \\
\hline
\end{tabular}%
}
\tightfloatafter
\end{table*}

\tightsection{Additional RQ4 Target-Exposure Details}
\label{sec:supp_rq4_details}

RQ4 uses UR5eUMI and GoogleRobot as representative target embodiments for the pretrain-exposed zero-shot setting.
UR5eUMI tests transfer to an arm with different degrees of freedom from the source embodiments, while GoogleRobot tests transfer to a mobile-manipulator platform with more humanoid-like morphology.
For each target, we construct a separate 640K-trajectory pick-and-place pretraining pool.
At each exposure ratio, a controlled fraction of non-target source data is replaced with target-embodiment pretraining data, so the total pretraining budget remains fixed.
The target embodiment's downstream post-training tasks are never used in these scaling experiments; thus the evaluation remains zero-shot with respect to target-embodiment downstream tasks, but differs from strict zero-shot transfer because the target embodiment appears during pretraining.

The target-embodiment post-training oracle is initialized from the same shared pretrained checkpoint as the zero-shot variants and subsequently post-trained on $\mathcal{D}(e_t,\mathcal{T}_{\mathrm{post}})$.
It is not a zero-shot setting and serves only as a performance upper reference.
In the RQ4 target-exposure plot, the pretrain-exposed curves remain below this oracle, indicating that target-embodiment pretraining mitigates but does not eliminate the need for target-task adaptation.

\tightsection{Model Training Details}
\label{sec:supp_model_training_details}
We use a Mixture-of-Transformers (MoT) backbone following $\pi_{0.5}$~\cite{physicalintelligence2025pi05}, initializing the vision-language stream from pretrained VLM weights and coupling it with a separate action expert through layer-wise attention.
The visual input contains the current observation image resized to $224\times224$, and proprioceptive inputs use a 4-step state history with 99th-percentile normalization for continuous states and actions.
The policy predicts a 12-step action chunk.
Action outputs are supervised with a flow-matching imitation loss; auxiliary co-training targets, when enabled, use cross-entropy loss with weight 0.25 relative to the imitation loss.
Pretraining uses batch size 320 on 16 H800 GPUs for 120K steps, and post-training uses batch size 160 on 8 H800 GPUs for 100K steps.

\tightsection{Policy Inference and Action Execution Details}
\label{sec:supp_inference_details}

At inference time, each request contains the language instruction $\ell$, two RGB observations (one wrist-mounted view and one third-person view), and a four-entry EEF state history.
Images are resized to $224\times224$.
We use the notation from the main-text state-action representation table: $b$ denotes the robot base frame, $e_t$ denotes the gripper frame at time $t$, $T_{x,y}$ denotes the transform from frame $x$ to frame $y$, and $\mathcal{V}(T)=[p,\operatorname{Euler}(R)]$ for $T=\begin{bmatrix}R&p\\0&1\end{bmatrix}$.
The gripper state or command is appended as a scalar.

The server interface can be written as a request--response map.
For each setting, the request specifies the language instruction, images, and one of the two EEF state histories, while the response is a 12-step action chunk,
\[
\hat{A}_{t:t+11}=\{\hat{a}_j\}_{j=t}^{t+11},
\qquad
\hat{a}_j=[\delta v_j,u_j]\in\mathbb{R}^{7},
\qquad
\delta v_j=[\Delta p_j,\Delta \phi_j]\in\mathbb{R}^{6}.
\]
Here $\Delta \phi_j$ uses the same Euler-angle convention as $\mathcal{V}(T)$, and the continuous gripper output $u_j$ is quantized to $\{-1,0,+1\}$, corresponding to close, no-op, and open.
The four request--response forms are:
\begin{enumerate}[leftmargin=*]
\item \textbf{Abs. EEF state / World-Delta action.}
The server request is
\[
\mathcal{Q}_t^{\mathrm{Abs}}=
\left(\ell,I_t^{\mathrm{wrist}},I_t^{\mathrm{third}},X_t^{\mathrm{Abs}}\right),
\qquad
X_t^{\mathrm{Abs}}=
\left\{
\left[\mathcal{V}\!\left(T_{b,e_{t-i}}\right),g_{t-i}\right]
\right\}_{i=0}^{3}.
\]
The server response is
\[
\mathcal{R}_t^{\mathrm{World}\text{-}\Delta}
=
\hat{A}_{t:t+11}^{\mathrm{World}\text{-}\Delta},
\qquad
\delta v_j=
\mathcal{V}(\bar{T}_{b,e_{j+1}})-\mathcal{V}(T_{b,e_j}),
\]
which gives the absolute gripper-frame target by
\[
\mathcal{V}(\bar{T}_{b,e_{j+1}})=\mathcal{V}(T_{b,e_j})+\delta v_j.
\]

\item \textbf{Abs. EEF state / EEF-Delta action.}
The server request is
\[
\mathcal{Q}_t^{\mathrm{Abs}}=
\left(\ell,I_t^{\mathrm{wrist}},I_t^{\mathrm{third}},X_t^{\mathrm{Abs}}\right),
\qquad
X_t^{\mathrm{Abs}}=
\left\{
\left[\mathcal{V}\!\left(T_{b,e_{t-i}}\right),g_{t-i}\right]
\right\}_{i=0}^{3}.
\]
The server response is
\[
\mathcal{R}_t^{\mathrm{EEF}\text{-}\Delta}
=
\hat{A}_{t:t+11}^{\mathrm{EEF}\text{-}\Delta},
\qquad
\delta v_j=\mathcal{V}(\Delta T_j),
\qquad
\Delta T_j=T_{b,e_j}^{-1}\bar{T}_{b,e_{j+1}}.
\]
Writing $\delta v_j=[\Delta p_j,\Delta\phi_j]$, the response is converted to an absolute gripper-frame target as
\[
\Delta T_j=
\begin{bmatrix}
\operatorname{Euler}^{-1}(\Delta\phi_j) & \Delta p_j\\
0 & 1
\end{bmatrix},
\qquad
\bar{T}_{b,e_{j+1}}=T_{b,e_j}\Delta T_j .
\]

\item \textbf{EEF-Delta state / World-Delta action.}
The server request is
\[
\mathcal{Q}_t^{\mathrm{EEF}\text{-}\Delta}=
\left(\ell,I_t^{\mathrm{wrist}},I_t^{\mathrm{third}},X_t^{\mathrm{EEF}\text{-}\Delta}\right),
\]
with
\[
X_t^{\mathrm{EEF}\text{-}\Delta}=
\left\{
\left[\mathcal{V}\!\left(T_{b,e_t}^{-1}T_{b,e_{t-i}}\right),g_{t-i}\right]
\right\}_{i=0}^{3}
=
\left\{
\left[\mathcal{V}\!\left(T_{e_t,e_{t-i}}\right),g_{t-i}\right]
\right\}_{i=0}^{3}.
\]
The server response is
\[
\mathcal{R}_t^{\mathrm{World}\text{-}\Delta}
=
\hat{A}_{t:t+11}^{\mathrm{World}\text{-}\Delta},
\qquad
\delta v_j=
\mathcal{V}(\bar{T}_{b,e_{j+1}})-\mathcal{V}(T_{b,e_j}),
\]
which gives the absolute gripper-frame target by
\[
\mathcal{V}(\bar{T}_{b,e_{j+1}})=\mathcal{V}(T_{b,e_j})+\delta v_j.
\]

\item \textbf{EEF-Delta state / EEF-Delta action.}
The server request is
\[
\mathcal{Q}_t^{\mathrm{EEF}\text{-}\Delta}=
\left(\ell,I_t^{\mathrm{wrist}},I_t^{\mathrm{third}},X_t^{\mathrm{EEF}\text{-}\Delta}\right),
\]
with
\[
X_t^{\mathrm{EEF}\text{-}\Delta}=
\left\{
\left[\mathcal{V}\!\left(T_{b,e_t}^{-1}T_{b,e_{t-i}}\right),g_{t-i}\right]
\right\}_{i=0}^{3}
=
\left\{
\left[\mathcal{V}\!\left(T_{e_t,e_{t-i}}\right),g_{t-i}\right]
\right\}_{i=0}^{3}.
\]
The server response is
\[
\mathcal{R}_t^{\mathrm{EEF}\text{-}\Delta}
=
\hat{A}_{t:t+11}^{\mathrm{EEF}\text{-}\Delta},
\qquad
\delta v_j=\mathcal{V}(\Delta T_j),
\qquad
\Delta T_j=T_{b,e_j}^{-1}\bar{T}_{b,e_{j+1}}.
\]
Writing $\delta v_j=[\Delta p_j,\Delta\phi_j]$, the response is converted to an absolute gripper-frame target as
\[
\Delta T_j=
\begin{bmatrix}
\operatorname{Euler}^{-1}(\Delta\phi_j) & \Delta p_j\\
0 & 1
\end{bmatrix},
\qquad
\bar{T}_{b,e_{j+1}}=T_{b,e_j}\Delta T_j .
\]
\end{enumerate}

For efficiency, the server batches up to 10 requests.
In our serving setup, one warmed-up policy query on a server with a single NVIDIA GeForce RTX 4090 GPU takes $168\,\mathrm{ms}$ and uses $6168\,\mathrm{MiB}$ of GPU memory.
% Original placeholder: \todowtxy{Describe the inference logic of both EEF-Delta representation here. Also details about inference speed and memory usage.}
% Original working-tree placeholder: r

\tightsection{Additional Problem-Setup Details}
\label{sec:supp_problem_details}

Each demonstration consists of observation--action pairs collected from embodiment $e$ performing task $t$.
Observations include images and language instructions, while proprioceptive states and actions are expressed in the representation specified by each experimental condition.
The policy $\pi_\theta$ maps the past observation history and task specification to an action chunk, $\pi_\theta: (o_{<t}, \ell) \mapsto a_{t:t+K}$, where $o_{<t}$ is the observation history before the current action step $t$, $\ell$ is the language instruction, and $a_{t:t+K}$ denotes an action chunk of length $K$.
We use $\mathcal{E}_{\mathrm{pre}}$ and $\mathcal{T}_{\mathrm{pre}}$ to denote the embodiments and tasks represented in $\mathcal{D}_{\mathrm{pre}}$.
Across comparisons, the policy backbone, data budget, optimization protocol, testing tasks, scenes, cameras, and objects are held fixed unless explicitly stated.

\end{document}